\documentclass[journal]{IEEEtran}

\usepackage[utf8]{inputenc}
\usepackage[T1]{fontenc} % T1 字体编码，支持欧洲语言字符
\usepackage{amsmath,amsfonts,amssymb} % AMS 数学宏包
\usepackage{amsthm}
\usepackage{bm}
\usepackage{newtxtext} % 用于设置正文为 Times 字体
\usepackage{newtxmath} % 用于设置数学公式为 Times 风格字体
\usepackage{graphicx}                % 用于插入图片
\usepackage{cite}                    % 改进引用命令
\usepackage{url}                     % 用于显示 URL

\usepackage{booktabs}                % 专业美观的表格线
\usepackage{multirow}                % 表格中的多行单元格
\usepackage{algorithm}               % 算法浮动环境
\usepackage[noend]{algpseudocode}    % 算法伪代码样式
\usepackage{subcaption}              % 子图环境
\usepackage{xcolor}                  % 颜色支持
\usepackage{enumitem}
\usepackage{tabularx}
\usepackage{etoolbox}
\usepackage{pifont}                  % 提供 dingbats 符号，包括 \ding

\newtheoremstyle{mydef_bolditalic}
  {\topsep}   % 上方间距
  {\topsep}   % 下方间距
  {\normalfont} % 正文内容字体 (标准正体)
  {1em}       % 段首缩进
  {\bfseries\itshape} % 标题字体 (粗斜体)
  {.}         % 标题后的标点
  {5pt plus 1pt minus 1pt} % 标题与正文的间距
  {\thmname{#1}\thmnumber{ #2}\thmnote{ (#3)}} % 标题格式

\theoremstyle{mydef_bolditalic}

\newtheorem{definition}{Definition}

\begin{document}
\onecolumn

% --- TITLE AND AUTHOR INFORMATION ---
\title{A Novel Approach to Temporal QoS Estimation via Extended Kalman Filter-Incorporated Latent Feature Analysis}

\author{
    Ye Yuan,~\IEEEmembership{Member,~IEEE,}
    Song Wang,
    Hongxun Zhou,
    Ling Wang,
    and Xin Luo,~\IEEEmembership{Fellow,~IEEE}
    
    \thanks{This research is supported in part by the National Key Research and Development Program of China under Grant 2024YFF0908200, in part by the National Natural Science Foundation of China under grant 62372385, and in part by the New Chongqing Youth Innovation Talent Projectand under Grant CSTB2025YITP-QCRCX0054 (Corresponding author: X. Luo).}
    \thanks{Ye. Yuan, Song Wang, Hongxun Zhou, and Xin. Luo are with the College of Computer and Information Science, Southwest University, Chongqing 400715, China (e-mail: yuanyekl@swu.edu.cn, plutocharonh@gmail.com, libnova7@swu.edu.cn, luoxin@ieee.org).}
    \thanks{L. Wang is with the School of Computer Science and Technology, Chongqing University of Posts and Telecommunications, Chongqing 400065, China (e-mail: wangling1820@gmail.com).}
}

% Paper headers
\markboth{IEEE Journal Template, Vol. XX, No. XX, September 2025}%
{Yuan \MakeLowercase{\textit{et al.}}: An Effective QoS Predictor via EKL}

\maketitle

% --- ABSTRACT AND KEYWORDS ---
\begin{abstract}
Predicting temporal Quality of Service (QoS) data is critical for optimizing network services and rationalizing resource allocation in cloud computing and service-oriented systems. Existing mainstream methods have achieved promising predictive performance. However, their purely data-driven manner limits their ability to capture non-stationary temporal patterns, thereby leading to accuracy degradation when temporal QoS data exhibits fluctuations. To tackle this limitation, we propose a novel \underline{E}xtended  \underline{K}alman Filter-Enhanced  \underline{L}atent Feature Analysis (EKL) model to perform efficient and accurate temporal QoS prediction from the perspective of bidirectional model–data–driven. Its main idea is three-fold: a) designing a model-driven feature producer to obtain the temporal latent features to capture the intricate temporal pattern following the principle of an Extended Kalman Filter; b) building a data-driven feature producer based on the alternating least squares algorithm to identify time-invariant latent features describing intrinsic user-service characteristics; c) exploiting a density-oriented parallel strategy that achieves workload balancing by sorting users in accordance with their service invocation density, which effectively elevates computational efficiency. In addition, we provide a rigorous theoretical analysis to formally prove the convergence of the proposed EKL. Experimental evaluations conducted on real-world temporal QoS datasets reveal that our proposed EKL surpasses existing state-of-the-art models with respect to both computational efficiency and prediction accuracy for missing temporal QoS data.

\end{abstract}


\IEEEpeerreviewmaketitle

% --- SECTION 1: INTRODUCTION ---
\section{Introduction}

\IEEEPARstart{D}{ue} to the rapid advancement of cloud computing, various service providers offer a multitude of functionally equivalent Web services \cite{adeleye2023constructing, purohit2023qos}. Hence, how to select the most appropriate services for users is a vital yet thorny issue \cite{liu2024qosgnn, lu2024qos}. Quality of Service (QoS) reflects the non-functional characteristics (e.g., response time and throughput) of Web services \cite{lu2024qos, ghahramani2017toward}, which plays an important role in service selection \cite{savasci2023ddpc, lu2024qos}. Note that QoS data is frequently acquired by the warming-up test \cite{cao2024framework}. However, it is invariably time-consuming and financially costly to evaluate all the candidate services \cite{kulshrestha2024transient}. Hence, building an effective QoS predictor is of significant importance for service selection \cite{chai2024emd}.

Generally, the QoS data can be described by a user-service QoS matrix, whose element represents the QoS record for a specific metric experienced by a user on a service \cite{wu2023double, yuan2024adaptive, li2026adaptivepid, yuan2025pid, yuan2024fuzzypid}. Due to the potentially vast number of services, it becomes impractical for a user to access all the offered services. Hence, a user-service matrix is commonly incomplete containing numerous missing elements \cite{bi2023two, chen2024hyloref, li2025errorrefinement, yuan2023divergence}. Recently, a latent feature analysis (LFA) model is specially designed for an incomplete matrix and widely adopted to predict the missing QoS data owing to its simplicity and scalability \cite{wu2023double, bi2023two, he2019manifold, he2021learning, he2024polarized}. However, in the real scenario of cloud computing, the QoS data is commonly temporally varying \cite{li2024dynamic, bojovic2022dynamic, rayani2022ensuring, yuan2022multilayered, yuan2020generalized, bi2024fastautoencoder}. Hence, the temporal QoS data can be fully described by a sequence of user-service matrices that change over time. Unfortunately, the aforementioned latent factor analysis (LFA) models typically treat data as independent slices, thus overlooking inherent temporal patterns and sequential dependencies \cite{wu2024predictionsampling, wu2024mmlf, li2024nesterov, zhong2024admm}. Hence, how to design a competitive temporal QoS predictor has emerged as a high-priority research topic in recent years.

\begin{figure}[!htbp]
    \centering

    % --------- Subfigure (a) ---------
    \begin{subfigure}[b]{0.48\columnwidth} % 第一个子图，占据0.48列宽
        \centering
        \includegraphics[width=\textwidth]{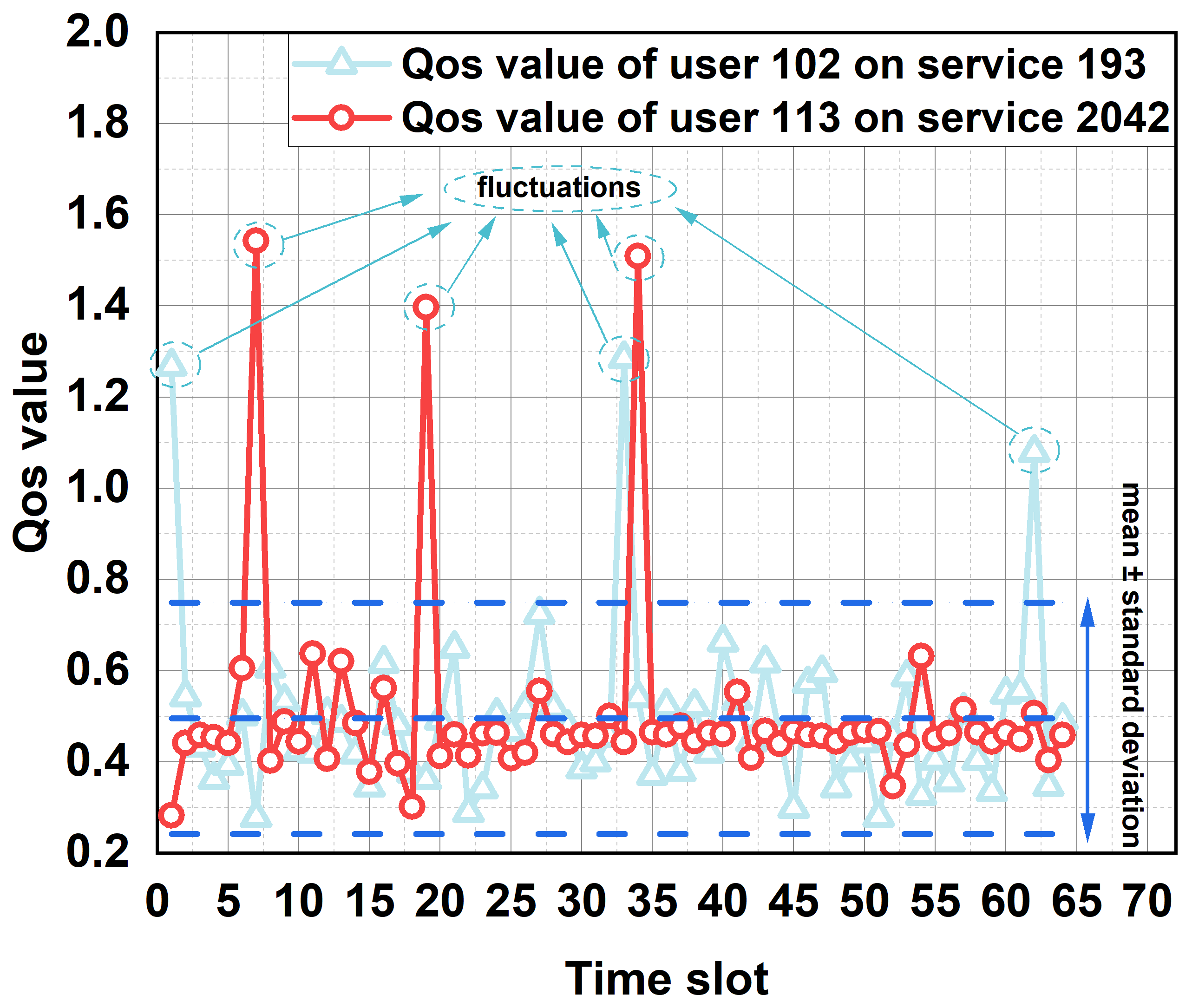} 
        \caption{} % 子图标题留空或按需修改
        \label{fig:fluctuation:a}
    \end{subfigure}
    \hfill % 在两个子图之间添加水平空间
    % --------- Subfigure (b) ---------
    \begin{subfigure}[b]{0.48\columnwidth} % 第二个子图，占据0.48列宽
        \centering
        \includegraphics[width=\textwidth]{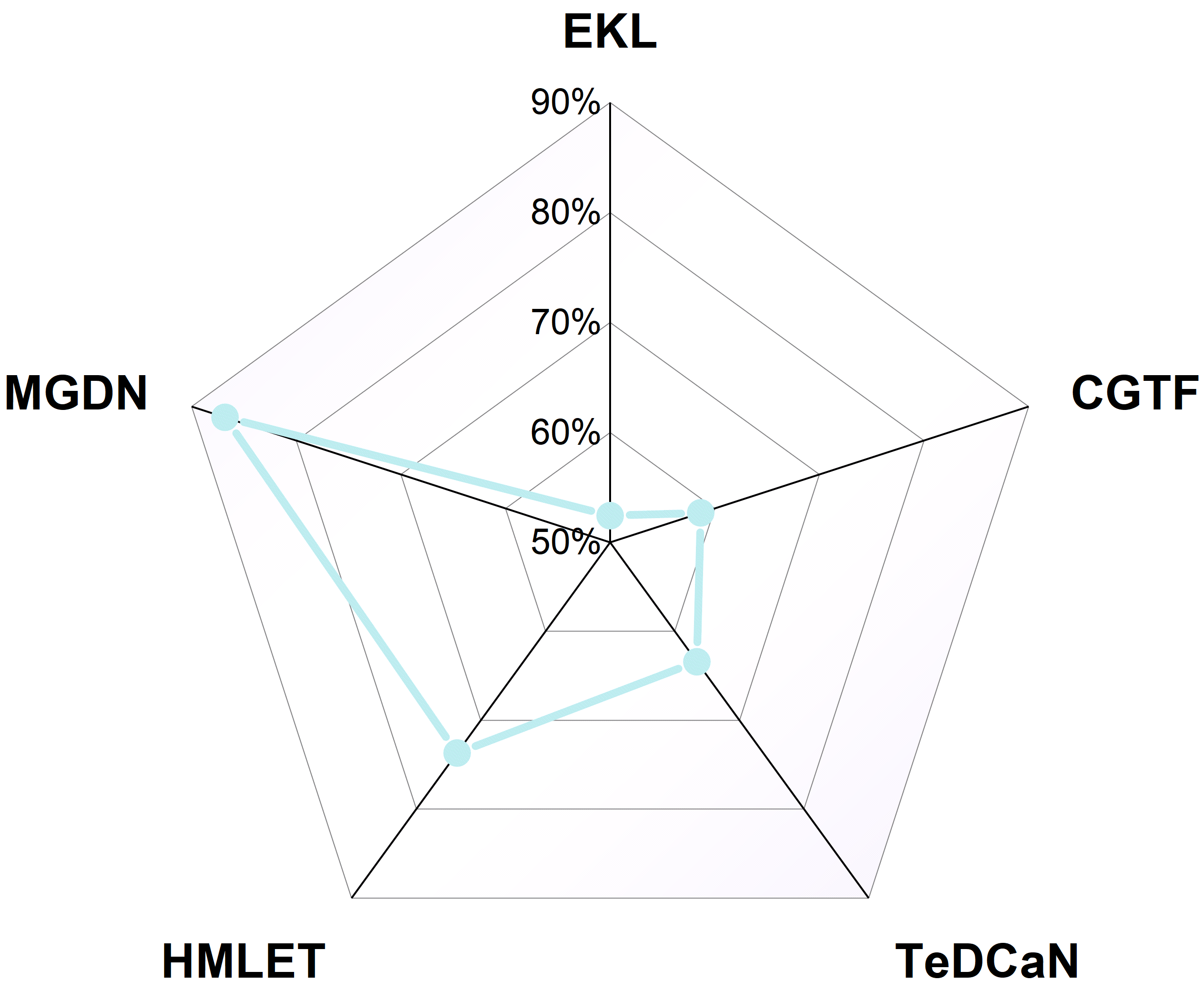} 
        \caption{} % 子图标题留空或按需修改
        \label{fig:fluctuation:b}
    \end{subfigure}
    
    % --------- Main Caption  ---------
    \caption{Prediction bias on fluctuations. We define the fluctuations based on the statistical characteristics of QoS data. Specifically, the baseline is the mean value of all QoS data and fluctuations  that lie outside the range of `mean ± standard deviation'. Fig. 1(a) shows that QoS data contains fluctuation components, which merely account for a fraction portion of the total QoS data. Moreover, Fig. 1(b) displays the deviation between RMSE on fluctuation components and its corresponding RMSE on all QoS data. This clearly shows that all the temporal QoS predictors exhibit varying predictive biases inherently. However, the proposed EKL possesses the lowest prediction bias compared with the existing data-driven models. Note that MGDN, CGTF, HMLET and TeDCaN are the compared models in experimental section.}
    \label{fig:fluctuation}
\end{figure}

For carrying out representation learning for such temporal QoS data, great efforts have been made \cite{luo2020temporal, tang2024temporal, xu2023time, che2024tensor, shin2024pgcn, zhu2023wingnn}. To date,  temporal LFA models like tensor factorization \cite{tang2024temporal, yuan2023kalman}, dynamic graph neural network \cite{shin2024pgcn, zeng2024tensordetector}, and dynamic matrix factorization \cite{yu2016temporal, wu2024finegrained} are the mainstream approaches since they adopt advanced strategies to capture the temporal patterns of QoS data. However, it is noteworthy that the QoS data is prone to fluctuate suddenly in real Web service scenarios due to unforeseeable factors, i.e., network issue, hardware failure, and human operational error \cite{shi2021service, xu2025adaptively, wang2025convolutionbias, liao2025proximaladmm}. Although the above temporal QoS predictors already obtain excellent predictive performance, they are all purely data-driven \cite{multi2024multi}. When processing the QoS data with fluctuations over time, their representation ability is still limited \cite{luo2020temporal, tang2024temporal, xu2023time, said2022spatiotemporal, yuan2020temporal, xu2026sampling}. As shown in Fig. 1, it is easy to see that the purely data-driven approaches are insensitive to fluctuation components. The main reason is that the purely data-driven manner typically depends on the latent pattern extracted from historical QoS data, which contains a substantial portion of steady varying components \cite{tian2023scan, zhou2022fedformer, chen2024nesterovtensor, yang2025temporalregularized}. Hence, these steady varying components far exceed that of fluctuation components, which may introduce the possibility of bias in QoS prediction. 

According to prior studies in the automation community \cite{yuan2023cyber, chang2023tensor}, an Extended Kalman Filter (EKF) is a nonlinear recursive filter model, which is widely adopted to capture the complex temporal patterns of non-stationary temporal data precisely from the perspective of a model-driven approach. Currently, it is widely used in various nonlinear temporal systems to track the varying state with the state-transition and observation functions in practical applications. For instance, Chang \textit{et al.} \cite{chang2023tensor} adopt EKF to track the inherent nonlinear temporal patterns in realistic temporal traffic data. Zhao \textit{et al.} \cite{zhao2017robust} estimate the transients of a power system in a faster and more reliable way based on EKF.

In this context, we assume that the temporal QoS data reflects the properties of both the user- and service-sides comprehensively. Specifically, user-side properties like packet loss and latency reflect intricate temporal patterns, \textit{i.e.}, temporal latent features \cite{luo2020temporal}. Moreover, service-side properties like communication protocol, price, functional configuration, and hardware specification correspond to intrinsic characteristics, i.e., time-invariant latent features \cite{bi2023two}. Hence, based on the above assumptions and investigations, this study proposes a novel \underline{EK}F-enhanced \underline{L}FA (EKL) model to perform highly efficient and accurate QoS prediction. Its main ideas are:
\begin{enumerate}[label=\alph*)]
    \item Designing a model-driven feature producer (MFP) to obtain the temporal latent features to capture the intricate temporal patterns following the principle of EKF;
    \item Building a data-driven feature producer (DFP) to identify the time-invariant latent features to describe the intrinsic characteristics based on alternating least squares (ALS) algorithm;
    \item Exploiting a density-oriented parallel strategy (DPS) to improve the computational efficiency.
\end{enumerate}

The main contributions of this study include:
\begin{enumerate}[label=\alph*)]
    \item \textbf{We propose a novel EKL model}. It builts a novel bidirectional model–data driven learning framework to accurately represent temporal QoS data. Specifically, it adopts a control model, i.e., EKF, to learn the temporal latent features, and utilize an ALS algorithm to explore the time-invariant latent features from the a data-driven perspective, which is the core novelty. Moreover, it equips a rapid computing capability by incorporating a density-oriented parallel strategy;
    \item \textbf{We conduct the rigorous convergence analysis}. It proves that the convergence of EKL is guaranteed with the optimization process relying on EKF and ALS;
    \item \textbf{We present the extensive experiments on real-world QoS datasets}. The outcomes of experimental results demonstrate that the proposed EKL delivers remarkable gains in accuracy and efficiency compared with state-of-the-art models for missing QoS data estimation \cite{wu2026nongradient}.
\end{enumerate}

\section{Related Work}

\subsection{{Tensor Factorization}}
Tensor factorization, i.e., Canonical Polyadic \cite{haliassos2022supervised, he2026tensorlowrank, qin2026electricity} and Tucker \cite{fang2024functional, tang2025autoencoding, tang2025neural} regard the temporal QoS data as a three-order tensor, and maps it into low-dimensional space, thereby constructing a desired low-rank approximation of QoS data subsequently. For instance, Tang \textit{et al.} \cite{tang2024temporal, wu2025modeaware, liao2025tensorcausal} propose a biased non-negative tensor factorization model to perform an accurate description on temporal QoS data. Che \textit{et al.} \cite{che2024tensor, li2026neuralnltf} propose a graph-regularized tensor factorization model, which captures the complex connections with a tensor. He \textit{et al.} \cite{he2022bayesian, chen2025battery} propose a Bayesian tensor factorization model to effectively interpolate the incomplete tensor with large observation missing. Ioannidis \textit{et al.} \cite{ioannidis2021coupled, xu2025attention} propose an innovative model termed coupled graph-tensor factorization that aptly incorporates graph-associated auxiliary information. Xu \textit{et al.} \cite{xu2023hrstlr} propose a Hessian regularization spatio-temporal low rank algorithm to extract the temporal and spatial correlations effectively. Bhanu \textit{et al.} \cite{bhanu2021embedding, hou2025multiaspect} propose a tensor decomposition method with characteristic network constraints that considers location-pair reciprocity for low-rank tensor approximation.

\subsection{{Dynamic Graph Neural Network}}
Dynamic graph neural network has shown potential in handling temporal QoS data. It mainly utilizes a static graph neural network \cite{das2024ags, wang2026graphtensor, wang2026highorder} and a sequence neural network \cite{pareja2020evolvegcn, han2025sgddyg, yuan2025nodecollab} to explore the spatial-temporal patterns within dynamic data. For instance, Shin \textit{et al.} \cite{shin2024pgcn, wang2025gta2t} propose a progressive dynamic graph neural network, which establishes stepwise adjacency matrices by capturing trend similarities among different nodes, and utilizes activation components to retrieve temporal features. Nazzal \textit{et al.} \cite{nazzal2024semi, bi2025dynamicgraphmixer} propose a heterogeneous graph neural network-LSTM algorithm to leverage multiple edge types. Zhou \textit{et al.} \cite{zhou2023spatial, he2025modularized} propose a context-aware dynamic graph neural network to measure the similarity of users or services based on time-varying QoS fluctuation. Yuan \textit{et al.} \cite{yuan2024dynamic, bi2025graphlinear} propose a robust by introducing the minimal-sufficient-consensual condition, thereby addressing the spatial-temporal information flow. Cini \textit{et al.} \cite{cini2023scalable} propose a scalability-enabled architecture capitalizing on high-efficiency spatiotemporal dynamics encoding.

\subsection{{Dynamic Matrix Factorization}}
Dynamic matrix factorization aims to incorporate temporal patterns into matrix factorization models, enabling effective representation learning for temporal data \cite{lyu2026dynamic, hu2025parallelreview}. For instance, Mohammadiha \textit{et al.} \cite{mohammadiha2014state, qin2024adaptiveparallel} propose a state-space dynamic nonnegative matrix factorization model that captures linear temporal trends. Yu \textit{et al.} \cite{yu2016temporal, qin2024asynchronous} build a temporal regularized matrix factorization model by embedding autoregressive dependencies to effectively handle temporal patterns. Bhavana \textit{et al.} \cite{bhavana2025temporal, lyu2025genetic} explore temporal evolutions of latent factors using polynomial functions to realize precise prediction. Koren \textit{et al.} \cite{koren2009collaborative, qin2024paralleladaptive} adopt time-varying biases into matrix factorization to model long-term user factors and short-term temporal shifts. Chatzis \textit{et al.} \cite{chatzis2014dynamic, wang2024distributed} propose a dynamic probabilistic matrix factorization model with Bayesian hierarchical priors to characterize the dynamic distributions of latent factors.

According to the above investigations, the proposed EKL differs from existing models in following perspectives:
\begin{enumerate}[label=\alph*)]
    \item It adopts the bidirectional model–data driven learning framework for temporal QoS estimation. In contrast, existing models are purely data-driven approaches;
    \item It learns the temporal user latent features and time-invariant service latent features, which is consistent with physical nature of QoS. However, existing models treat all the latent features as time-varying;
    \item It rigorously proves that EKL is ensured to converge in theory with the bidirectional model–data driven learning process, which is not implemented by existing studies.
\end{enumerate}

% --- SECTION 3: PRELIMINARIES ---
\section{Preliminaries}

\subsection{Symbol and Definition}
\begin{figure}[!htbp]
    \centering
    \includegraphics[width=\columnwidth]{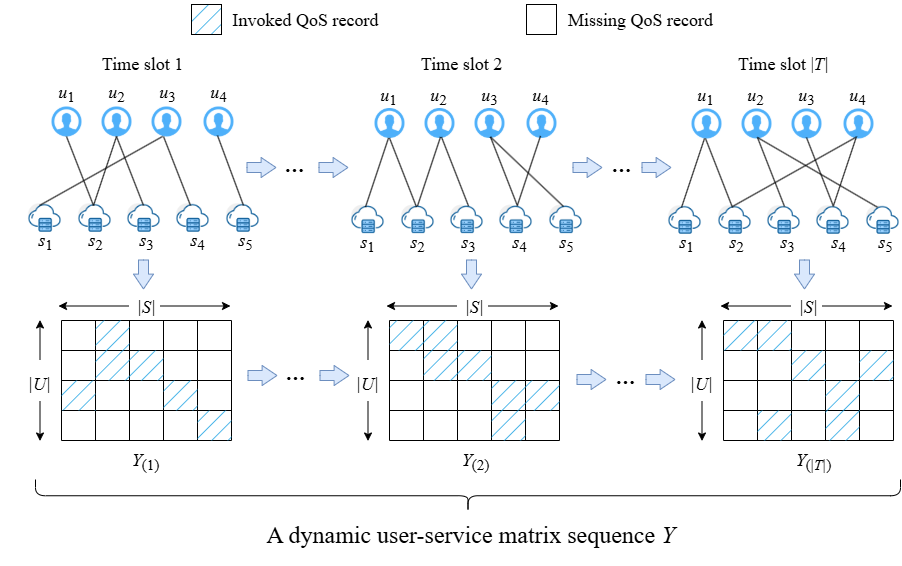} 
    \caption{An illustrative example of a temporal user-service matrix sequence.}
    \label{fig:matrix_sequence}
\end{figure}

The temporal QoS data can be comprehensively characterized by a temporal user-service matrix set, which is defined as follows:

\begin{definition}[\textbf{A Temporal User-Service Matrix Sequence}]
Given a user set $U$, service set $S$, and time slot set $T$, ${Y}^{|U|\times|S|\times|T|}=\{{Y}_{(1)}, {Y}_{(2)}, \dots, {Y}_{(|T|)}\}$ denotes a temporal user-service matrix sequence, as depicted in Fig.~\ref{fig:matrix_sequence}. Especially, ${Y}_{(t)}$ is a user-service matrix at time slot $t \in T$, and $y_{(t)u,s}$ is the element of ${Y}_{(t)}$ to describe a QoS record of invocation by $u \in U$ on $s \in S$. Note that it is impossible for a user to invoke all candidate services. Therefore, ${Y}$ is commonly incomplete and $\Lambda$ denotes its known entry sets.
\end{definition}

\begin{definition}[\textbf{Temporal Latent and Time-Invariant Latent Features}]
Given a latent feature dimension $f$, ${P}^{|U|\times|f|\times|T|}=\{{P}_{(1)}, {P}_{(2)}, \dots, {P}_{(|T|)}\}$ denotes the temporal latent features, which reflects the temporal preference of users related with time. Moreover, ${Q}^{|S|\times|f|}$ denotes the time-invariant latent features which captures intrinsic stable characteristics and is consistent across all time slots.
\end{definition}

An EKL-based QoS predictor is defined as follows:

\begin{definition}[\textbf{An EKL-based QoS Predictor}]
Given a temporal user-service matrix sequence ${Y}$, EKL learns ${P}$ and ${Q}$ to build its approximation $\hat{{Y}}$ based on $\Lambda = \sum_{t=1}^{|T|} \Lambda(t)$. Concretely, it minimizes the loss $\sum_{t=1}^{|T|} \sum_{y_{(t)u,s} \in \Lambda(t)} (y_{(t)u,s} - p_{(t)u} (q_s)^T)^2$ to generate the predicted QoS value $\hat{y}_{(t){u,s}} = {p}_{(t)u} ({q}_s)^T$, where ${p}_{(t)u}$ is the row vector of ${P}$ related to user $u$ at time $t$ and ${q}_s$ is the row vector of ${Q}$ related to service $s$.
\end{definition}

\subsection{Extended Kalman Filter}
The Extended Kalman Filter (EKF) is a recursive nonlinear filtering model designed to accurately predict the evolving states of a non-stationary temporal system. Specifically, an EKF leverages state-transition and observation functions to capture complex temporal patterns in a non-stationary temporal system as follows:
\begin{subequations}
\label{eq:ekf_base}
\begin{align}
    x_t &= S(x_{t-1}) + w_{t-1}, \label{eq:ekf_state} \\
    z_t &= O(x_t) + r_t, \label{eq:ekf_obs}
\end{align}
\end{subequations}
where $x_t$ is the state vector of time slot $t$, $z_t$ is the observation vector, $S(\cdot)$ is the nonlinear activation state-transition function, $O(\cdot)$ is the nonlinear activation observation function, $w_{t-1}$ and $r_t$ is the state-transition and observation Gaussian noise with zero mean and covariance $C[w_{t-1}]$ and $C[r_t]$.

% --- SECTION 4: The Proposed EKL-based QoS Predictor ---
\section{The Proposed EKL-based QoS Predictor}
This section is dedicated to presenting the comprehensive details of our proposed EKL, which performs an accurate representation of the temporal QoS data in a bidirectional model-data driven manner. It consists of the following three modules:
\begin{enumerate}[label=\alph*)]
    \item A model-driven feature producer (MFP) learns temporal latent features under the guidance of the EKF principle;
    \item A data-driven feature producer (DFP) learns time-invariant latent features based on the principle of ALS;
    \item A density-oriented parallel strategy (DPS) refines the computational efficiency.
\end{enumerate}

\subsection{Model-Driven Feature Producer}
In order to capture the complex temporal patterns existing within a temporal user-service matrix sequence describing the temporal QoS data, we initially describe the temporal latent features ${P}$ following the state-transition and observation functions \cite{liu2024symmetry, wu2025multimetric, wu2025robustlowrank, wu2025outlier, li2025knowledge}. Specifically, for a specific user $u \in U$, the state-transition function is built to represent the dependencies between the state ${p}_{(t)u}$ and ${p}_{(t-1)u}$ at adjacent time slots:
\begin{equation}
    p_{(t)u} = S(p_{(t-1)u}) + w_{(t-1)u},
    \label{eq:state_transition_p}
\end{equation}
where ${p}_{(t)u}$ denotes the row vector of ${P}$ related to user $u$ at time slot $t$, ${w}_{(t-1)u}$ denotes the state-transition noise with Gaussian distribution ${N}(0, C[{w}_{(t-1)u}])$, and $S({p}_{(t-1)u})$ describes the nonlinear state variation from ${p}_{(t-1)u}$ to ${p}_{(t)u}$.

Further, we utilize the observation function to establish the relationship between the temporal latent features ${P}$ with the invoked QoS records. In accordance with the principle of EKF, we formulate the subsequent observation process equation:
\begin{equation}
    \label{eq:observation_base}
    y_{(t)u} = O(p_{(t)u}) + r_{(t)u},
\end{equation}
where $y_{(t)u}$ is the observation data, which denotes the invoked QoS records of user $u$ at time slot $t$, $r_{(t)u}$ denotes the observation noise with Gaussian distribution ${N}(0, C[r_{(t)u}])$, and $O(p_{(t)u})$ is the nonlinear activation observation function.

As shown in \textbf{\textit{Definition 3}}, the invoked QoS records are decided by temporal latent and time-invariant latent features simultaneously, i.e., $p_{(t)u}(q_s)^T$. Hence, we reformulated (3) into the following form to accurately map the current state $p_{(t)u}$ to the observation space:
\begin{equation}
    y_{(t)u} = O(p_{(t)u})(M_{(t)u})^T + r_{(t)u},
    \label{eq:obs_reformulated}
\end{equation}
where $M_{(t)u}$ is the subset of $Q$ and denotes the service’s time-invariant latent feature set invoked by user $u$ at time slot $t$. Specifically, for a specific user $u$, the construction process of $M_{(t)u}$ is shown in Fig.~\ref{fig:mtu_construction}. In our context, we define these two nonlinear activation functions $S(\cdot)$ and $O(\cdot)$ as LeakyReLU, i.e., $\max(\alpha x, x)$, where $\alpha$ is a small positive constant like 0.01. The main reason is that LeakyReLU provides a satisfied tradeoff between expressiveness, stability, and efficiency for state tracking.

Note that for each user $u \in U$, we build an EKF to characterize $\{p_{(1)u}, p_{(2)u}, \dots, p_{(|T|)u}\}$ based on \eqref{eq:state_transition_p} and \eqref{eq:obs_reformulated}. Hence, $|U|$ independent EKFs are obtained to exploit the temporal latent features $P=\{P_{(1)}, P_{(2)}, \dots, P_{(|T|)}\}$. Following the principle of EKF, we obtain the desired $P$ through two steps:

\begin{figure}[!htbp]
    \centering
    \includegraphics[width=\columnwidth]{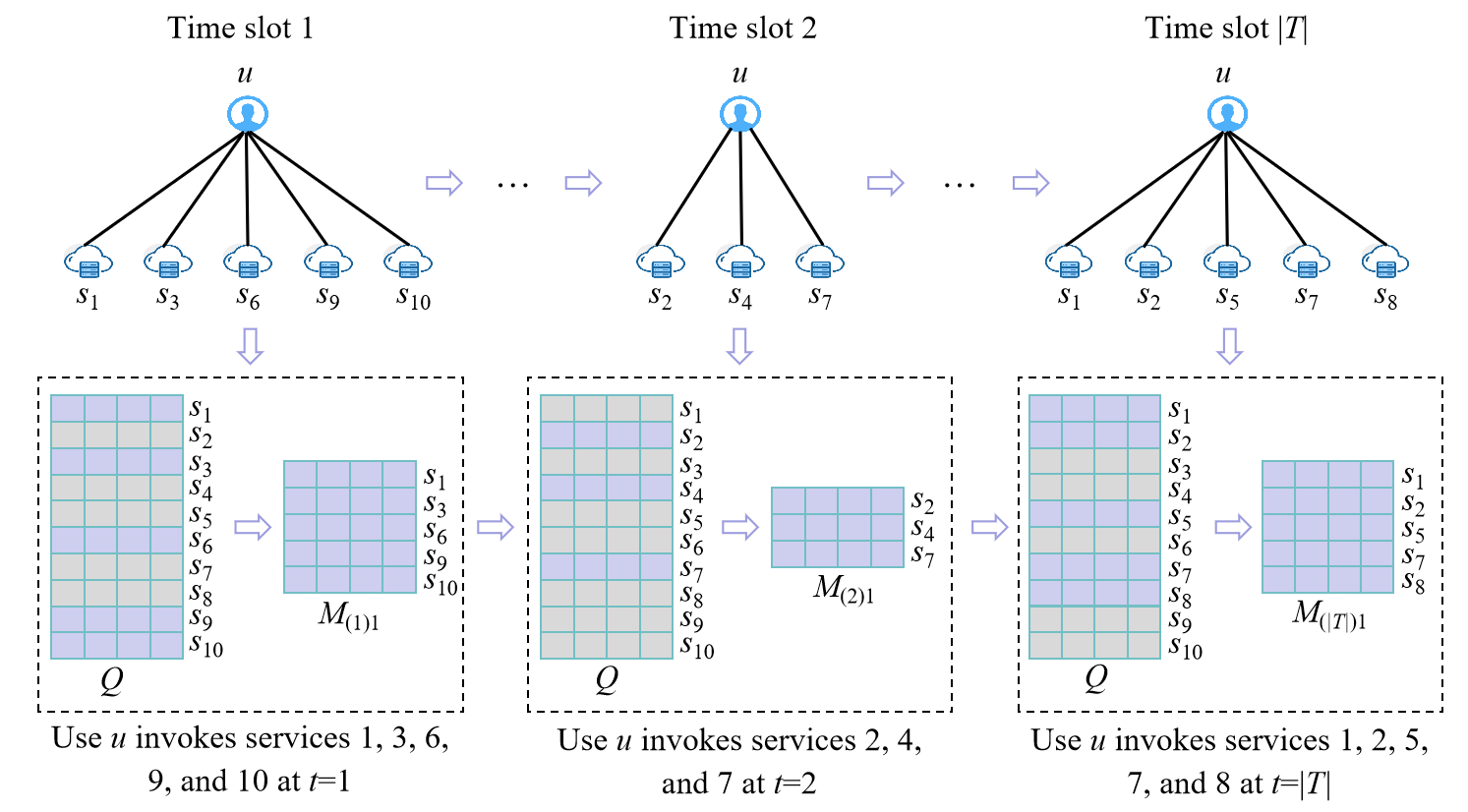} % Placeholder for Fig. 3
    \caption{The construction process of ${M}_{(t)u}$.}
    \label{fig:mtu_construction}
\end{figure}

\noindent % 防止段首缩进
\textbf{{a) Prediction Step:}}

This step utilizes the state-transition function (2) to predict the state at next time slot:
\begin{equation}
    p_{-(t)u} = S(p_{(t-1)u}),
    \label{eq:predict_state}
\end{equation}
\begin{equation}
    C[p_{-(t)u}] = S'(p_{(t)u}) C[p_{(t-1)u}] (S'(p_{(t)u}))^T + C[w_{(t-1)u}],
    \label{eq:predict_cov}
\end{equation}
where $p_{-(t)u}$ is the prior prediction of $p_{(t)u}$, $S'(p_{(t-1)u})$ is the partial derivative of $S(p_{(t-1)u})$. $C[p_{-(t)u}]$, $C[p_{(t-1)u}]$ and $C[w_{(t-1)u}]$ denotes the covariance matrix of $p_{-(t)u}$, $p_{(t-1)u}$ and $w_{(t-1)u}$, respectively.

\noindent % 防止段首缩进
\textbf{b) Update Step:} 

This step mainly utilizes the observation data to refine the prior prediction $p_{-(t)u}$ at the time slot $t$, thereby obtaining the final state $p_{(t)u}$. Specifically, the specific update process is shown as follows:
% --- Formula (7) ---
\begin{equation}
    \label{eq:update_gain}
    K_{(t)u} = \frac{C[p_{-(t)u}](D(p_{(t)u}))^T}{(D(p_{(t)u})C[p_{-(t)u}](D(p_{(t)u}))^T + C[r_{(t)u}])},
\end{equation}
% --- Formula (8) ---
\begin{equation}
    \label{eq:update_state}
    p_{(t)u} = p_{-(t)u} + K_{(t)u}(y_{(t)u} - O(p_{-(t)u})(M_{(t)u})^T),
\end{equation}
% --- Formula (9) ---
\begin{equation}
    \label{eq:update_cov}
    C[p_{(t)u}] = C[p_{(t-1)u}] - K_{(t)u}D(p_{(t)u})C[p_{(t-1)u}],
\end{equation}
% --- Accompanying Text ---
where $D(p_{(t)u})=O(p_{(t)u})(M_{(t)u})^T$, $K_{(t)u}$ is the gain matrix. For $\forall u \in U$, $P=\{P_{(1)}, P_{(2)}, \dots, P_{(T)}\}$ is obtained based on \eqref{eq:predict_state}-\eqref{eq:update_cov}.

\subsection{Data-Driven Feature Producer}
After obtaining the temporal latent features $P=\{P_{(1)}, P_{(2)}, \dots, P_{(|T|)}\}$, we solve the time-invariant latent features $Q$ alternately. Specifically, the data-driven optimization algorithm ALS \cite{hou2023progressive} is adopted to achieve this goal. As depicted in \textbf{\textit{Definition 3}}, the loss function is
\begin{equation}
    \label{eq:loss_base}
    \varepsilon(P,Q) = \sum_{t=1}^{|T|} \sum_{y_{(t)u,s} \in \Lambda_{(t)}} (y_{(t)u,s} - p_{(t)u} (q_s)^T)^2.
\end{equation}

In order to prevent overfitting, $L_2$ regularization is adopted. Moreover, $P$ is acquired based on EKF and subsequently alternately anchored for further processing. Hence, the overall objective function can be reformulated:
\begin{equation}
    \label{eq:loss_regularized}
    \varepsilon(Q) = \sum_{t=1}^{|T|} \sum_{y_{(t)u,s} \in \Lambda_{(t)}} \left( \lambda(y_{(t)u,s} - p_{(t)u} (q_s)^T)^2 + \|q_s\|^2_2 \right),
\end{equation}
where $\lambda$ denotes the regularization coefficient.

Note that for $\forall s \in S$, the partial loss of \eqref{eq:loss_regularized} is given:
\begin{equation}
    \label{eq:loss_partial}
    \varepsilon(q_s) = \sum_{t=1}^{|T|} \sum_{y_{(t)u,s} \in \Lambda_{(t)s}} \left( \lambda(y_{(t)u,s} - p_{(t)u} (q_s)^T)^2 + \|q_s\|^2_2 \right),
\end{equation}
where $\Lambda_{(t)s}$ is the subset of $\Lambda_{(t)}$ corresponding to service $s$. Further, we divide \eqref{eq:loss_partial} into the following linear equations to solve it conveniently:
\begin{equation}
\label{eq:loss_derivative}
\varepsilon(q_s) =
\begin{cases}
\lambda (y_{(1)u,s} - p_{(1)u}(q_s)^T)^2 + \|q_s\|^2, & \forall u \in \Lambda_{(1)s} \\
\lambda (y_{(2)u,s} - p_{(2)u}(q_s)^T)^2 + \|q_s\|^2, & \forall u \in \Lambda_{(2)s} \\
\vdots & \vdots \\
\lambda (y_{(|T|)u,s} - p_{(|T|)u}(q_s)^T)^2 + \|q_s\|^2, & \forall u \in \Lambda_{(|T|)s}
\end{cases}
\end{equation}

\begin{figure}[!htbp]
    \centering
    \includegraphics[width=\columnwidth]{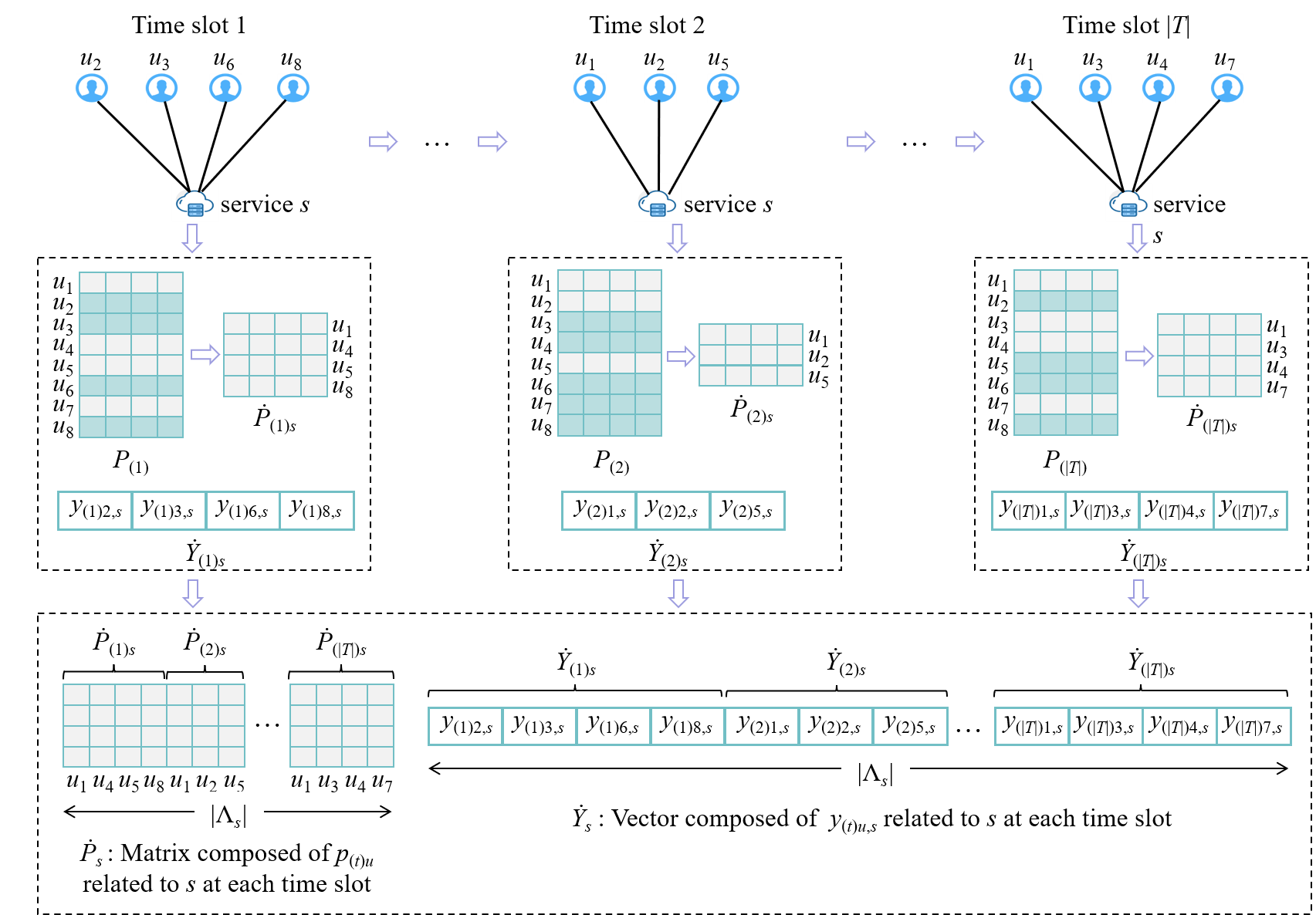} % Placeholder for Fig. 4
    \caption{Detailed construction processes of $\dot{{Y}}_s, \dot{{P}}_s$, and $|\Lambda_s|$.}
    \label{fig:construction_Y_P}
\end{figure}

Therefore, \eqref{eq:loss_derivative} can be further merged into:
\begin{equation}
    \label{eq:loss_merged}
    \varepsilon(q_s) = \lambda \|\dot{Y}_s - q_s \dot{P}_s \|^2 + |\Lambda_s| \|q_s\|^2,
\end{equation}
where $\dot{Y}_s$ is the vector composed of the invoked QoS records $y_{(t)u,s}$ related to $s$ at each time slot, $\dot{P}_s$ is the matrix composed of $p_{(t)u}$ related to $s$ at each time slot, and $|\Lambda_s|$ is the length of $\dot{Y}_s$. For instance, considering a specific service $s$, users $u_2, u_3, u_6$ and $u_8$ invoke it at $t=1$, users $u_1, u_2$ and $u_5$ invoke it at $t=2$, and users $u_1, u_3, u_4$ and $u_7$ invoke it at $t=|T|$. That means different user sets invoke a specific service at different times. Hence, their detailed construction processes are depicted in Fig.~\ref{fig:construction_Y_P}.

Based on \eqref{eq:loss_merged}, $q_s$ can be solved following the principle of ALS algorithm as follows:
\begin{equation}
    \label{eq:solve_q}
    \begin{split}
        \frac{\partial \varepsilon(q_s)}{\partial q_s} &= -\lambda(\dot{Y}_s - q_s \dot{P}_s)(\dot{P}_s)^T + |\Lambda_s|q_s = 0 \\
        \Rightarrow q_s &= \dot{Y}_s(\dot{P}_s)^T \left( \frac{|\Lambda_s|}{\lambda}I + \dot{P}_s(\dot{P}_s)^T \right),
    \end{split}
\end{equation}
where $I$ denotes the identity matrix. Note that the time-invariant latent features $Q$ can be obtained by employing the learning scheme \eqref{eq:solve_q} for all the $s \in S$.

\subsection{{Density-Oriented Parallel Strategy}}
From \eqref{eq:predict_state}-\eqref{eq:update_cov}, it is clear to see that the generation of temporal latent features $P$ involves a large number of matrix operations. Hence, it is essential to improve the computational efficiency. Fortunately, for $\forall u \in U$, their corresponding EKFs are independent of each other. Therefore, it is natural to consider using multi-threaded parallel computing to improve computational efficiency of the MFP.

Note that the temporal QoS data is highly incomplete. Hence, it is easy to see that different users invoke different numbers of services in our context. For a specific user $u \in U$, the number of invoked QoS records are shown as follows:
\begin{equation}
    \label{eq:num_invoked}
    |\Lambda_u| = |y_{(1)u}| + |y_{(2)u}| + \dots + |y_{(|T|)u}|,
\end{equation}
where $\Lambda_u$ denotes the QoS set invoked by user $u$, whose number varies significantly corresponding to different users. Given the variability in invoked QoS record volume across users, a straightforward implementation of multi-threaded parallel computing causes an uneven workload distribution, thereby further leading to threads waiting for each other.

\begin{figure}[!htbp]
    \centering
    \includegraphics[width=\columnwidth]{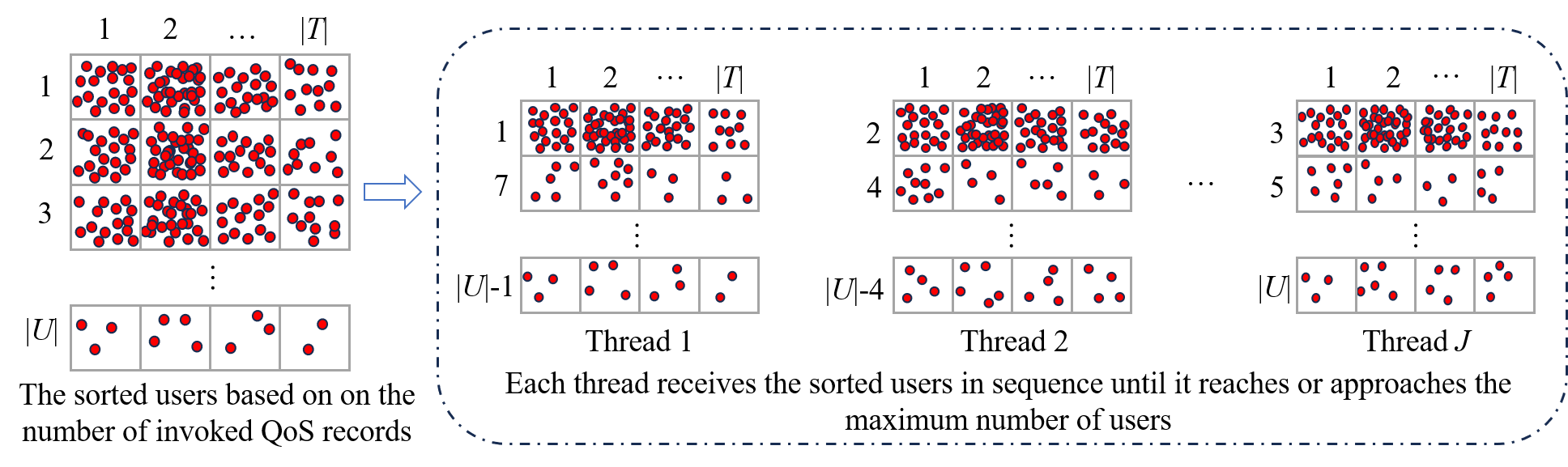} % Placeholder for Fig. 5
    \caption{The density-oriented parallel strategy. The red points denote the invoked QoS record volume with user $u$.}
    \label{fig:dps}
\end{figure}

To mitigate it, a simple yet effective density-oriented parallel strategy is proposed, as shown in Fig.~\ref{fig:dps}. Firstly, it ensures the number of users assigned to each thread is roughly identical. Given the number of thread $J$, the maximum number of users per thread is calculated as follows:
\begin{equation}
    \label{eq:max_users_thread}
    M = \lceil |U|/|J| \rceil,
\end{equation}
where $\lceil \cdot \rceil$ rounds up the result to the nearest integer.

Secondly, all the users are sorted based on the number of invoked QoS records. Specifically, we rank the users in descending order, and then reverse the ranking from the middle segment. It results in an overall order that presents a pattern of descending followed by ascending. Finally, it assigns the sorted users to each thread. Specifically, users are assigned to threads in a round‑robin manner according to the sorted order until each thread reaches the maximum number of users, thereby significantly enhancing the computational efficiency.

%%%%%%%%%%%%%%%%%%%%%%%%%%%%%%%%%%%%%%%%%%%%%%%%%%%%%%%%%%%%
%  SECTION 5: EXPERIMENTS (COMPLETE)
%%%%%%%%%%%%%%%%%%%%%%%%%%%%%%%%%%%%%%%%%%%%%%%%%%%%%%%%%%%%

\section{Experiments} 

\subsection{General Settings}

\textbf{Datasets.} Three real-world QoS datasets are adopted in the experiments, including two from the WS-DREAM project \cite{luo2022novel} (Response Time and Throughput) and invocations between microservices from Alibaba production clusters\cite{luo2021characterizing}. Their details are summarized in Table~\ref{tab:datasets}. For each dataset, we design four testing cases with varying training-validation-testing ratios to evaluate model performance under different data sparsity levels, as depicted in Table~\ref{tab:test_cases}.

\begin{table}[!h]
\centering
\caption{Details of datasets.}
\label{tab:datasets}
\begin{tabular}{lccc}
\toprule
\textbf{Dataset}          & \textbf{D1} & \textbf{D2} & \textbf{D3} \\ \midrule
Name & Throughput  & Response Time &Alibaba    \\
User Count        & 142         & 142           & 1000         \\
Service Count     & 4500        & 4500          & 15000        \\
Time Slot Count   & 64          & 64            & 144          \\
Time Slot Length      &15 mins     &15 mins       &5 mins      \\  
Total Time Covered     &16 hours    &16 hours      &12 hours     \\  
Known Entry Set   & 30,287,611  & 30,287,611    &  1,892,000 \\ \bottomrule
\end{tabular}
\end{table}

\begin{table}[!h]
\centering
\caption{Detailed testing cases.}
\label{tab:test_cases}
\begin{tabular}{lll}
\toprule
\textbf{No.} & \textbf{Cases} & \textbf{Training-Validation-Testing} \\ \midrule
\multirow{4}{*}{D1} & D11 & 3\%-12\%-85\% \\
                    & D12 & 5\%-20\%-75\% \\
                    & D13 & 10\%-40\%-50\% \\
                    & D14 & 15\%-60\%-25\% \\ \midrule
\multirow{4}{*}{D2} & D21 & 3\%-12\%-85\% \\
                    & D22 & 5\%-20\%-75\% \\
                    & D23 & 10\%-40\%-50\% \\
                    & D24 & 15\%-60\%-25\% \\ \midrule
\multirow{4}{*}{D3} & D31 & 3\%-12\%-85\% \\
                    & D32 & 5\%-20\%-75\% \\
                    & D33 & 10\%-40\%-50\% \\
                    & D34 & 15\%-60\%-25\% \\ \bottomrule
\end{tabular}
\end{table}

\textbf{Baselines.} To verify the efficacy of the proposed EKL, twelve state-of-the-art temporal LFA models are adopted in our study, i.e., CGTF\cite{ioannidis2021coupled}, TeDCaN\cite{bhanu2021embedding}, HRST-LR\cite{xu2023hrstlr}, HMLET\cite{kong2022linear}, GTN\cite{yun2019graph}, SGP\cite{cini2023scalable}, MGDN\cite{hu2024mgdcf}, PGCN\cite{shin2024pgcn}, WD-GCN\cite{manessi2020dynamic}, SGL-ED\cite{wu2021self}, hetGNN-LSTM\cite{nazzal2024semi}, TM-GCN\cite{malik2021dynamic} and SDNMF\cite{mohammadiha2014state}.

% --- 方案三代码（最终推荐版） ---

\textbf{Evaluation Metrics.} Missing QoS estimation accuracy is mainly considered to evaluate the performance. The commonly used evaluation metrics are root mean square error (RMSE) and mean absolute error (MAE) \cite{luo2020temporal, wu2023double}. The smaller RMSE/MAE denotes the higher estimation accuracy. Generally, they are formulated as:
\begin{align*}
    \text{RMSE} &= \sqrt{ \left( \sum_{(u,s,t) \in Y_{\Phi}} (y_{(t)u,s} - \hat{y}_{(t)u,s})^2 \right) \Big/ |Y_{\Phi}| }, \\
    \text{MAE} &= \sqrt{ \left( \sum_{(u,s,t) \in Y_{\Phi}} |y_{(t)u,s} - \hat{y}_{(t)u,s}|_{\text{abs}} \right) \Big/ |Y_{\Phi}| },
\end{align*}
where $Y_{\Phi}$ stands for the testing dataset, and $|\cdot|_{\text{abs}}$ is used to calculate the absolute value of the enclosed numerical value, respectively. Empirical experiments are carried out on a computing node featuring an Intel(R) Xeon(R) Gold 5218 CPU, GeForce RTX 3050 GPU, and 512GB RAM.

\textbf{Configuration.} In order to maintain objective results, the following conditions are adopted: a) the termination criteria is that estimation accuracy deteriorates continuously for 10 epochs or the iterations reaches the threshold, i.e., 1000; b) for each testing case, we repeat ten times to achieve the final averaged output; c) for EKL, the latent feature dimension is fixed as 10, and the number of threads is fixed as 16 when compared with other baselines; and d) for baselines, we follow the parameter settings recommended in their original papers and tune them via grid search to obtain the optimal results.

\subsection{Comparison Performance}

This section presents a comparative analysis of the proposed EKL with several state-of-the-art baselines. From these results, we have the following findings:
\begin{enumerate}[label=\alph*)]
    \item \textbf{EKL’s accuracy is higher than that of its peers when estimating the missing QoS.} As shown in Table~\ref{tab:results_accuracy_full}, the proposed EKL obtains the optimal estimation results on 13 cases out of total 16 cases. For instance, on D13, EKL obtains the lowest RMSE at 0.1667, which is 5.66\% lower than 0.1767 of CGTF, 2.69\% lower than 0.1713 of TeDCaN, 47.49\% lower than 0.3175 of HRST-LR, 28.58\% lower than 0.2334 of HMLET, 44.19\% lower than 0.2987 of GTN, 66.94\% lower than 0.5042 of SGP, 46.28\% lower than 0.3103 of MGDN, 65.52\% lower than 0.4835 of PGCN, 66.84\% lower than 0.5027 of WD-GCN, 42.75\% lower than 0.2912 of SGL-ED, 64.65\% lower than 0.4716 of hetGNN-LSTM, 67.07\% lower than 0.5063 of TM-GCN, and 6.71\% lower than 0.1787 of SDNMF. The main reason is that the proposed EKL performs highly efficient and accurate QoS prediction from the perspective of bidirectional model-data driven. However, EKL occasionally exhibits slightly lower prediction accuracy than individual baseline in specific cases, i.e., TeDCaN in D11/RMSE, D21/RMSE, and D22/RMSE. This may be because M3 utilize tensor decomposition to explore temporal characteristics as well as adopt graph Laplacian constraint to engrave the topological structure of user-service interactions. It is should be pointed out that SDNMF is a state-space-based dynamic matrix factorization model. However, it is still outperformed by EKL. The reason is that SDNMF utilize an EM algorithm to obtain the temporal latent features from a purely data-driven manner. However, EKL builts a bidirectional model-data driven learning framework, which obtains the temporal latent features based EKF, and utilizes an ALS algorithm to explore the time-invariant latent features from a data-driven perspective. Moreover, SDNMF learns the temporal patterns based on a linear state-space. In contrast, EKL adopts a nonlinear activation to improve the representation ability.
    
    \item \textbf{EKL exhibits excellent computational efficiency when learning the temporal QoS data.} As reported in Table~\ref{tab:results_efficiency}, the proposed EKL achieves the lowest total time cost in 12 out of 16 test cases. For instance, on D11, EKL consumes 51 seconds to reach the lowest MAE, which is 0.36\% of CGTF's 13982 seconds, 0.28\% of TeDCaN' 18273 seconds, 1.48\% of HRST-LR' 3440 seconds, 0.06\% of HMLET' 81153 seconds, 0.08\% of GTN' 63840 seconds, 0.43\% of SGP' 11776 seconds, 0.38\% of MGDN' 13455 seconds, 0.11\% of PGCN' 45239 seconds, 0.86\% of WD-GCN' 5915 seconds, 0.18\% of SGL-ED' 29049 seconds, 0.25\% of hetGNN-LSTM' 20705 seconds, 0.08\% of TM-GCN' 60165 seconds, and 48.57\% of SDNMF' 105 seconds. It is mainly because that EKL adopts the density-oriented parallel strategy to improve the computational efficiency. Moreover, EKL only relies on the known set of temporal QoS data.
    
    \item \textbf{The performance improvement of EKL is statistical significance.} To comprehensively evaluate EKL’s performance, we employ the Wilcoxon test and Friedman test. The Friedman test is used to validate the performance of multiple models across multiple datasets, and the Wilcoxon signed-ranks test is adopted to conduct a pairwise performance difference analysis between EKL and each baseline. As shown in Tables~\ref{tab:results_accuracy_full} and~\ref{tab:results_efficiency}, EKL obtains the lowest Rank value, which indicates that it acquires the optimal performance in both estimation accuracy and computational efficiency. In addition, the \textit{p}-values are much lower than the significance level of 0.05, which denote that EKL outperforms other comparative models.

    \item Moreover, we have additionally conducted some interesting experiments. Firstly, we select CGTF and TeDCaN as the baselines and implement Exponential Moving Average (EMA) on the latent factors. The results are recorded in the Table S1 of the Supplementary File. From it, the performance of EMA-CGTF and EMA-TeDCaN shows no improvement or even a slight degradation. This demonstrates that the performance gain of EKL stems specifically from the EKF rather than just generic smoothing. Fruther, We have added comprehensive experiments on fluctuation samples. The results in Table S2 of the Supplementary File show that EKL outperforms all baselines, which indicates its strong ability to process sudden QoS fluctuations. Finally, we simulate EKL's ability to process concept drift. Specifically, we split the dataset into three parts, using the first 70\% of time slices as the training set, the middle 20\% as the validation set, and the last 10\% as the test set. The results are shown in Table S3 of the Supplementary File. Under the temporal split setting, EKL still maintains competitive performance. The main reason is that the state-transition function of EKL possesses an extrapolation capability.  
       
\end{enumerate}

% 请确保在文档开头有 \usepackage{bm}

% =============================================================================
%                             TABLE V (REORDERED ONLY)
% =============================================================================
% 请确保导言区有：
% \usepackage[table]{xcolor}
% \usepackage{array}
% \usepackage{multirow}
% \usepackage{makecell}

% ==================== Table 1: Accuracy (Estimation Error) ====================
\begin{table*}[!htbp]
\centering
\caption{The comparison results on estimation error (RMSE/MAE), where $\star$ indicates that EKL is outperformed by the compared models.}
\label{tab:results_accuracy_full}
\setlength{\tabcolsep}{1.2pt} 
\resizebox{\textwidth}{!}{ 
% 最后的一列 c 更改为 >{\color{blue}}c 自动将该列内容标蓝
\begin{tabular}{llcccccccccccccc}
\toprule
\textbf{Case} & \textbf{Metric} & \textbf{EKL} & \textbf{CGTF} & \textbf{TeDCaN} & \textbf{HRST-LR} & \textbf{HMLET} & \textbf{GTN} & \textbf{SGP} & \textbf{MGDN} & \textbf{PGCN} & \textbf{WD-GCN} & \textbf{SGL-ED} & \textbf{hetGNN-LSTM} & \textbf{TM-GCN} & \textbf{SDNMF} \\ \midrule

\multirow{2}{*}{D11} & RMSE & 0.1786\textsubscript{$\pm$9E-4} & 0.1821\textsubscript{$\pm$3E-4} & $\star$ \textbf{0.1776\textsubscript{$\pm$3E-4}} & 0.3512\textsubscript{$\pm$6E-4} & 0.3278\textsubscript{$\pm$5E-4} & 0.3588\textsubscript{$\pm$1E-6} & 0.5042\textsubscript{$\pm$3E-4} & 0.3229\textsubscript{$\pm$2E-5} & 0.4845\textsubscript{$\pm$5E-4} & 0.5040\textsubscript{$\pm$3E-5} & 0.3696\textsubscript{$\pm$1E-4} & 0.4917\textsubscript{$\pm$9E-4} & 0.5060\textsubscript{$\pm$5E-3} & 0.1921\textsubscript{$\pm$4E-04} \\
                     & MAE  & \textbf{0.1053\textsubscript{$\pm$7E-4}} & 0.1106\textsubscript{$\pm$2E-4} & 0.1061\textsubscript{$\pm$6E-4} & 0.2649\textsubscript{$\pm$5E-4} & 0.2095\textsubscript{$\pm$3E-4} & 0.2410\textsubscript{$\pm$1E-6} & 0.3747\textsubscript{$\pm$3E-4} & 0.2195\textsubscript{$\pm$3E-5} & 0.3548\textsubscript{$\pm$4E-4} & 0.3774\textsubscript{$\pm$2E-5} & 0.2463\textsubscript{$\pm$1E-4} & 0.3642\textsubscript{$\pm$9E-4} & 0.3788\textsubscript{$\pm$5E-3} & 0.1135\textsubscript{$\pm$3E-04} \\ \midrule
\multirow{2}{*}{D12} & RMSE & \textbf{0.1717\textsubscript{$\pm$3E-4}} & 0.1814\textsubscript{$\pm$9E-4} & 0.1737\textsubscript{$\pm$3E-4} & 0.3178\textsubscript{$\pm$5E-4} & 0.2663\textsubscript{$\pm$2E-4} & 0.3170\textsubscript{$\pm$1E-4} & 0.5041\textsubscript{$\pm$4E-4} & 0.2815\textsubscript{$\pm$2E-4} & 0.4854\textsubscript{$\pm$2E-3} & 0.5036\textsubscript{$\pm$8E-4} & 0.3161\textsubscript{$\pm$3E-5} & 0.4792\textsubscript{$\pm$2E-2} & 0.5059\textsubscript{$\pm$1E-4} & 0.1827\textsubscript{$\pm$2E-04} \\
                     & MAE  & \textbf{0.1012\textsubscript{$\pm$2E-4}} & 0.1098\textsubscript{$\pm$1E-4} & 0.1026\textsubscript{$\pm$2E-4} & 0.2252\textsubscript{$\pm$4E-4} & 0.1740\textsubscript{$\pm$2E-4} & 0.2209\textsubscript{$\pm$2E-4} & 0.3759\textsubscript{$\pm$4E-4} & 0.1981\textsubscript{$\pm$2E-4} & 0.3525\textsubscript{$\pm$2E-3} & 0.3786\textsubscript{$\pm$1E-3} & 0.2213\textsubscript{$\pm$2E-5} & 0.3516\textsubscript{$\pm$2E-2} & 0.3782\textsubscript{$\pm$1E-4} & 0.1077\textsubscript{$\pm$2E-04} \\ \midrule
\multirow{2}{*}{D13} & RMSE & \textbf{0.1667\textsubscript{$\pm$5E-4}} & 0.1767\textsubscript{$\pm$3E-4} & 0.1713\textsubscript{$\pm$4E-4} & 0.3175\textsubscript{$\pm$4E-4} & 0.2334\textsubscript{$\pm$2E-4} & 0.2987\textsubscript{$\pm$6E-4} & 0.5042\textsubscript{$\pm$3E-4} & 0.3103\textsubscript{$\pm$1E-5} & 0.4835\textsubscript{$\pm$2E-3} & 0.5027\textsubscript{$\pm$1E-3} & 0.2912\textsubscript{$\pm$3E-5} & 0.4716\textsubscript{$\pm$7E-3} & 0.5063\textsubscript{$\pm$1E-4} & 0.1787\textsubscript{$\pm$4E-04} \\
                     & MAE  & \textbf{0.0982\textsubscript{$\pm$2E-4}} & 0.1061\textsubscript{$\pm$3E-4} & 0.1007\textsubscript{$\pm$2E-4} & 0.2240\textsubscript{$\pm$4E-4} & 0.1524\textsubscript{$\pm$1E-4} & 0.2148\textsubscript{$\pm$7E-4} & 0.3751\textsubscript{$\pm$4E-4} & 0.1819\textsubscript{$\pm$5E-4} & 0.3528\textsubscript{$\pm$1E-3} & 0.3766\textsubscript{$\pm$1E-3} & 0.2108\textsubscript{$\pm$3E-5} & 0.3489\textsubscript{$\pm$1E-3} & 0.3791\textsubscript{$\pm$1E-4} & 0.1048\textsubscript{$\pm$3E-04} \\ \midrule
\multirow{2}{*}{D14} & RMSE & \textbf{0.1651\textsubscript{$\pm$4E-4}} & 0.1739\textsubscript{$\pm$3E-3} & 0.1707\textsubscript{$\pm$3E-4} & 0.3172\textsubscript{$\pm$2E-4} & 0.2147\textsubscript{$\pm$2E-4} & 0.2956\textsubscript{$\pm$0E-8} & 0.5042\textsubscript{$\pm$3E-4} & 0.2831\textsubscript{$\pm$1E-5} & 0.4819\textsubscript{$\pm$4E-3} & 0.5044\textsubscript{$\pm$2E-4} & 0.2874\textsubscript{$\pm$3E-5} & 0.4664\textsubscript{$\pm$3E-3} & 0.5064\textsubscript{$\pm$1E-5} & 0.1808\textsubscript{$\pm$3E-04} \\
                     & MAE  & \textbf{0.0970\textsubscript{$\pm$2E-4}} & 0.1025\textsubscript{$\pm$2E-3} & 0.0999\textsubscript{$\pm$2E-4} & 0.2237\textsubscript{$\pm$2E-4} & 0.1355\textsubscript{$\pm$1E-4} & 0.2139\textsubscript{$\pm$1E-7} & 0.3603\textsubscript{$\pm$2E-2} & 0.1785\textsubscript{$\pm$1E-5} & 0.3534\textsubscript{$\pm$2E-3} & 0.3783\textsubscript{$\pm$1E-4} & 0.2093\textsubscript{$\pm$3E-5} & 0.3399\textsubscript{$\pm$6E-3} & 0.3799\textsubscript{$\pm$1E-5} & 0.1064\textsubscript{$\pm$2E-04} \\ \midrule

% --- Dataset D2 ---
\multirow{2}{*}{D21} & RMSE & 0.2056\textsubscript{$\pm$5E-4} & 0.2061\textsubscript{$\pm$2E-3} & $\star$ \textbf{0.2021\textsubscript{$\pm$4E-4}} & 0.2427\textsubscript{$\pm$1E-4} & 0.3128\textsubscript{$\pm$4E-4} & 0.3222\textsubscript{$\pm$0E-8} & 0.4118\textsubscript{$\pm$1E-4} & 0.2959\textsubscript{$\pm$7E-5} & 0.3880\textsubscript{$\pm$9E-4} & 0.4116\textsubscript{$\pm$3E-5} & 0.3526\textsubscript{$\pm$4E-5} & 0.3803\textsubscript{$\pm$7E-4} & 0.4140\textsubscript{$\pm$7E-5} & 0.2177\textsubscript{$\pm$5E-04} \\
                     & MAE  & \textbf{0.1150\textsubscript{$\pm$5E-4}} & 0.1190\textsubscript{$\pm$2E-3} & 0.1173\textsubscript{$\pm$7E-4} & 0.1504\textsubscript{$\pm$2E-4} & 0.1983\textsubscript{$\pm$2E-4} & 0.2129\textsubscript{$\pm$0E-8} & 0.3020\textsubscript{$\pm$7E-3} & 0.1924\textsubscript{$\pm$6E-5} & 0.2890\textsubscript{$\pm$1E-4} & 0.3090\textsubscript{$\pm$3E-4} & 0.2283\textsubscript{$\pm$2E-5} & 0.2787\textsubscript{$\pm$2E-3} & 0.3113\textsubscript{$\pm$3E-5} & 0.1230\textsubscript{$\pm$4E-04} \\ \midrule
\multirow{2}{*}{D22} & RMSE & 0.1984\textsubscript{$\pm$5E-4} & 0.2026\textsubscript{$\pm$3E-3} & $\star$ \textbf{0.1977\textsubscript{$\pm$7E-4}} & 0.2400\textsubscript{$\pm$2E-4} & 0.2815\textsubscript{$\pm$8E-4} & 0.3035\textsubscript{$\pm$1E-7} & 0.4117\textsubscript{$\pm$1E-4} & 0.2628\textsubscript{$\pm$9E-6} & 0.3884\textsubscript{$\pm$7E-4} & 0.4115\textsubscript{$\pm$7E-4} & 0.3135\textsubscript{$\pm$4E-5} & 0.3772\textsubscript{$\pm$1E-3} & 0.4136\textsubscript{$\pm$3E-5} & 0.2113\textsubscript{$\pm$4E-04} \\
                     & MAE  & \textbf{0.1111\textsubscript{$\pm$5E-4}} & 0.1177\textsubscript{$\pm$3E-3} & 0.1135\textsubscript{$\pm$7E-4} & 0.1476\textsubscript{$\pm$4E-4} & 0.1751\textsubscript{$\pm$7E-4} & 0.2078\textsubscript{$\pm$0E-8} & 0.3032\textsubscript{$\pm$8E-2} & 0.1761\textsubscript{$\pm$9E-6} & 0.2879\textsubscript{$\pm$7E-4} & 0.3068\textsubscript{$\pm$7E-4} & 0.2157\textsubscript{$\pm$3E-5} & 0.2771\textsubscript{$\pm$2E-2} & 0.3125\textsubscript{$\pm$3E-5} & 0.1171\textsubscript{$\pm$3E-04} \\ \midrule
\multirow{2}{*}{D23} & RMSE & \textbf{0.1817\textsubscript{$\pm$6E-4}} & 0.1981\textsubscript{$\pm$8E-4} & 0.1951\textsubscript{$\pm$2E-4} & 0.2295\textsubscript{$\pm$3E-4} & 0.2537\textsubscript{$\pm$5E-4} & 0.2932\textsubscript{$\pm$1E-4} & 0.4118\textsubscript{$\pm$6E-5} & 0.2455\textsubscript{$\pm$1E-5} & 0.3884\textsubscript{$\pm$1E-3} & 0.4116\textsubscript{$\pm$5E-4} & 0.2990\textsubscript{$\pm$4E-5} & 0.3619\textsubscript{$\pm$2E-2} & 0.4135\textsubscript{$\pm$3E-4} & 0.1991\textsubscript{$\pm$5E-04} \\
                     & MAE  & \textbf{0.1070\textsubscript{$\pm$5E-4}} & 0.1135\textsubscript{$\pm$1E-3} & 0.1112\textsubscript{$\pm$2E-4} & 0.1381\textsubscript{$\pm$1E-4} & 0.1500\textsubscript{$\pm$1E-4} & 0.2124\textsubscript{$\pm$1E-4} & 0.2940\textsubscript{$\pm$1E-2} & 0.1647\textsubscript{$\pm$9E-5} & 0.2886\textsubscript{$\pm$1E-3} & 0.3024\textsubscript{$\pm$1E-4} & 0.2172\textsubscript{$\pm$4E-5} & 0.2552\textsubscript{$\pm$2E-2} & 0.3135\textsubscript{$\pm$3E-4} & 0.1138\textsubscript{$\pm$4E-04} \\ \midrule
\multirow{2}{*}{D24} & RMSE & \textbf{0.1780\textsubscript{$\pm$9E-4}} & 0.1963\textsubscript{$\pm$1E-3} & 0.1930\textsubscript{$\pm$2E-4} & 0.2296\textsubscript{$\pm$4E-4} & 0.2358\textsubscript{$\pm$3E-4} & 0.2919\textsubscript{$\pm$6E-5} & 0.4119\textsubscript{$\pm$7E-5} & 0.2404\textsubscript{$\pm$1E-5} & 0.3859\textsubscript{$\pm$6E-4} & 0.4116\textsubscript{$\pm$2E-5} & 0.2612\textsubscript{$\pm$4E-5} & 0.3903\textsubscript{$\pm$2E-2} & 0.4124\textsubscript{$\pm$5E-4} & 0.1973\textsubscript{$\pm$6E-04} \\
                     & MAE  & \textbf{0.1061\textsubscript{$\pm$5E-4}} & 0.1113\textsubscript{$\pm$1E-3} & 0.1091\textsubscript{$\pm$2E-4} & 0.1383\textsubscript{$\pm$3E-4} & 0.1378\textsubscript{$\pm$2E-4} & 0.2147\textsubscript{$\pm$6E-5} & 0.2969\textsubscript{$\pm$1E-3} & 0.1540\textsubscript{$\pm$1E-5} & 0.2842\textsubscript{$\pm$6E-4} & 0.3094\textsubscript{$\pm$3E-5} & 0.1555\textsubscript{$\pm$2E-5} & 0.2930\textsubscript{$\pm$2E-2} & 0.3130\textsubscript{$\pm$5E-4} & 0.1108\textsubscript{$\pm$5E-04} \\ \midrule

% --- Dataset D3 ---
\multirow{2}{*}{D31} & RMSE & \textbf{0.1845\textsubscript{$\pm$2E-03}} & 0.2014\textsubscript{$\pm$2E-03} & 0.1966\textsubscript{$\pm$2E-03} & 0.1951\textsubscript{$\pm$7E-05} & 0.2654\textsubscript{$\pm$4E-05} & 0.6560\textsubscript{$\pm$1E-04} & 0.4520\textsubscript{$\pm$5E-03} & 0.6777\textsubscript{$\pm$8E-07} & 0.4245\textsubscript{$\pm$4E-03} & 0.4052\textsubscript{$\pm$1E-02} & 0.6778\textsubscript{$\pm$2E-03} & 0.4117\textsubscript{$\pm$2E-03} & 0.4718\textsubscript{$\pm$6E-04} & 0.2111\textsubscript{$\pm$3E-04} \\
                     & MAE  & \textbf{0.0847\textsubscript{$\pm$7E-04}} & 0.1176\textsubscript{$\pm$1E-03} & 0.1007\textsubscript{$\pm$9E-04} & 0.1030\textsubscript{$\pm$1E-04} & 0.1536\textsubscript{$\pm$9E-05} & 0.4901\textsubscript{$\pm$1E-04} & 0.3134\textsubscript{$\pm$2E-02} & 0.5285\textsubscript{$\pm$2E-06} & 0.2762\textsubscript{$\pm$4E-05} & 0.2635\textsubscript{$\pm$6E-3} & 0.5288\textsubscript{$\pm$2E-05} & 0.2706\textsubscript{$\pm$8E-03} & 0.2916\textsubscript{$\pm$8E-04} & 0.1043\textsubscript{$\pm$2E-04} \\ \midrule
\multirow{2}{*}{D32} & RMSE & \textbf{0.1602\textsubscript{$\pm$8E-04}} & 0.1879\textsubscript{$\pm$1E-03} & 0.1686\textsubscript{$\pm$9E-04} & 0.1927\textsubscript{$\pm$6E-05} & 0.2610\textsubscript{$\pm$2E-04} & 0.6236\textsubscript{$\pm$7E-04} & 0.5077\textsubscript{$\pm$6E-02} & 0.6771\textsubscript{$\pm$1E-05} & 0.4247\textsubscript{$\pm$5E-03} & 0.4110\textsubscript{$\pm$4E-03} & 0.6774\textsubscript{$\pm$1E-04} & 0.4070\textsubscript{$\pm$2E-03} & 0.4868\textsubscript{$\pm$6E-04} & 0.1764\textsubscript{$\pm$4E-04} \\
                     & MAE  & \textbf{0.0723\textsubscript{$\pm$4E-04}} & 0.1087\textsubscript{$\pm$8E-04} & 0.0892\textsubscript{$\pm$6E-04} & 0.1012\textsubscript{$\pm$3E-04} & 0.1520\textsubscript{$\pm$1E-04} & 0.4512\textsubscript{$\pm$5E-04} & 0.3639\textsubscript{$\pm$6E-02} & 0.5276\textsubscript{$\pm$3E-05} & 0.2812\textsubscript{$\pm$4E-05} & 0.2778\textsubscript{$\pm$5E-04} & 0.5284\textsubscript{$\pm$3E-03} & 0.2662\textsubscript{$\pm$6E-04} & 0.2954\textsubscript{$\pm$6E-04} & 0.0856\textsubscript{$\pm$3E-04} \\ \midrule
\multirow{2}{*}{D33} & RMSE & \textbf{0.1402\textsubscript{$\pm$7E-04}} & 0.1776\textsubscript{$\pm$9E-04} & 0.1616\textsubscript{$\pm$7E-04} & 0.1907\textsubscript{$\pm$1E-04} & 0.2590\textsubscript{$\pm$5E-05} & 0.5306\textsubscript{$\pm$1E-03} & 0.4295\textsubscript{$\pm$4E-03} & 0.6724\textsubscript{$\pm$4E-04} & 0.4135\textsubscript{$\pm$5E-04} & 0.4122\textsubscript{$\pm$3E-03} & 0.6678\textsubscript{$\pm$3E-04} & 0.3920\textsubscript{$\pm$2E-02} & 0.5057\textsubscript{$\pm$6E-04} & 0.1501\textsubscript{$\pm$5E-04} \\
                     & MAE  & \textbf{0.0644\textsubscript{$\pm$5E-04}} & 0.1035\textsubscript{$\pm$6E-04} & 0.0876\textsubscript{$\pm$4E-04} & 0.0999\textsubscript{$\pm$3E-04} & 0.1504\textsubscript{$\pm$1E-04} & 0.3581\textsubscript{$\pm$1E-03} & 0.2924\textsubscript{$\pm$1E-02} & 0.5171\textsubscript{$\pm$5E-04} & 0.2723\textsubscript{$\pm$3E-04} & 0.2708\textsubscript{$\pm$2E-03} & 0.5156\textsubscript{$\pm$3E-04} & 0.2502\textsubscript{$\pm$1E-02} & 0.3113\textsubscript{$\pm$7E-04} & 0.0720\textsubscript{$\pm$4E-04} \\ \midrule
\multirow{2}{*}{D34} & RMSE & \textbf{0.1335\textsubscript{$\pm$3E-04}} & 0.1625\textsubscript{$\pm$7E-04} & 0.1549\textsubscript{$\pm$5E-04} & 0.1902\textsubscript{$\pm$6E-05} & 0.2590\textsubscript{$\pm$4E-04} & 0.4678\textsubscript{$\pm$3E-04} & 0.4329\textsubscript{$\pm$7E-03} & 0.6549\textsubscript{$\pm$2E-04} & 0.4155\textsubscript{$\pm$5E-04} & 0.4069\textsubscript{$\pm$8E-03} & 0.6502\textsubscript{$\pm$2E-04} & 0.3637\textsubscript{$\pm$4E-03} & 0.5153\textsubscript{$\pm$6E-04} & 0.1421\textsubscript{$\pm$4E-04} \\
                     & MAE  & \textbf{0.0621\textsubscript{$\pm$3E-04}} & 0.0945\textsubscript{$\pm$5E-04} & 0.0728\textsubscript{$\pm$3E-04} & 0.0999\textsubscript{$\pm$2E-04} & 0.1503\textsubscript{$\pm$7E-06} & 0.2969\textsubscript{$\pm$2E-04} & 0.2977\textsubscript{$\pm$9E-03} & 0.4939\textsubscript{$\pm$2E-04} & 0.2728\textsubscript{$\pm$4E-04} & 0.2707\textsubscript{$\pm$7E-03} & 0.4951\textsubscript{$\pm$3E-04} & 0.2272\textsubscript{$\pm$5E-03} & 0.3198\textsubscript{$\pm$8E-04} & 0.0679\textsubscript{$\pm$3E-04} \\ \midrule

\multicolumn{2}{l}{\textbf{Win/Loss}} & -- & 24/0 & 21/3 & 24/0 & 24/0 & 24/0 & 24/0 & 24/0 & 24/0 & 24/0 & 24/0 & 24/0 & 24/0 & 24/0 \\ \midrule
% 使用 \color{blue} 使整行数字变蓝
\multicolumn{2}{l}{\textbf{Rank}}     & \textbf{1.12} & 3.62 & 2.12 & 6.00 & 5.83 & 9.17 & 11.79 & 8.67 & 10.17 & 11.00 & 9.92 &9.25 & 12.92 & 3.42 \\ \midrule
\multicolumn{2}{l}{\textbf{\textit{p}-value}}  & -- & \textbf{1.19E-07} & \textbf{1.63E-05} & \textbf{1.19E-07} & \textbf{1.19E-07} & \textbf{1.19E-07} & \textbf{1.19E-07} & \textbf{1.19E-07} & \textbf{1.19E-07} & \textbf{1.19E-07} & \textbf{1.19E-07} & \textbf{1.19E-07} & \textbf{1.19E-07} & \textbf{1.19E-07} \\ \bottomrule
\end{tabular}
}
\end{table*}

% ==================== Table 2: Efficiency (Computational Time) ====================
\begin{table*}[!htbp]
\centering
\caption{The comparison results on computational efficiency (seconds), where $\star$ indicates that EKL is outperformed by the compared models.}
\label{tab:results_efficiency}
\setlength{\tabcolsep}{1.2pt} 
\resizebox{\textwidth}{!}{ 
% 最后的一列 c 更改为 >{\color{blue}}c 自动将该列内容标蓝
\begin{tabular}{llcccccccccccccc}
\toprule
\textbf{Case} & \textbf{Metric} & \textbf{EKL} & \textbf{CGTF} & \textbf{TeDCaN} & \textbf{HRST-LR} & \textbf{HMLET} & \textbf{GTN} & \textbf{SGP} & \textbf{MGDN} & \textbf{PGCN} & \textbf{WD-GCN} & \textbf{SGL-ED} & \textbf{hetGNN-LSTM} & \textbf{TM-GCN} & \textbf{SDNMF} \\ \midrule

\multirow{2}{*}{D11} & RMSE & $\mathbf{46 \pm 3}$ & $14370 \pm 333$ & $18273 \pm 395$ & $3902 \pm 35$ & $80476 \pm 4504$ & $62731 \pm 3242$ & $12043 \pm 1433$ & $14296 \pm 6449$ & $51612 \pm 11367$ & $7083 \pm 664$ & $27515 \pm 477$ & $21754 \pm 1582$ & $75217 \pm 5426$ & $95 \pm 6$ \\
                     & MAE  & $\mathbf{51 \pm 4}$ & $13982 \pm 314$ & $18273 \pm 395$ & $3440 \pm 31$ & $81153 \pm 3438$ & $63840 \pm 2226$ & $11776 \pm 1055$ & $13455 \pm 6065$ & $45239 \pm 8672$ & $5915 \pm 551$ & $29049 \pm 336$ & $20705 \pm 1671$ & $60165 \pm 2033$ & $105 \pm 8$ \\ \midrule
\multirow{2}{*}{D12} & RMSE & $\mathbf{203 \pm 15}$ & $8848 \pm 646$ & $18570 \pm 143$ & $12738 \pm 52$ & $70236 \pm 6317$ & $53616 \pm 3342$ & $19812 \pm 2211$ & $14966 \pm 734$ & $57472 \pm 7719$ & $10930 \pm 965$ & $26235 \pm 372$ & $25592 \pm 1993$ & $69773 \pm 1684$ & $372 \pm 18$ \\
                     & MAE  & $\mathbf{215 \pm 15}$ & $9520 \pm 784$ & $18570 \pm 143$ & $11295 \pm 47$ & $71471 \pm 4377$ & $53542 \pm 2218$ & $13011 \pm 1405$ & $14034 \pm 682$ & $50239 \pm 7398$ & $12718 \pm 872$ & $26632 \pm 381$ & $24542 \pm 2110$ & $53694 \pm 1594$ & $413 \pm 26$ \\ \midrule
\multirow{2}{*}{D13} & RMSE & $\mathbf{1957 \pm 327}$ & $4209 \pm 711$ & $23800 \pm 2048$ & $15522 \pm 61$ & $62375 \pm 4128$ & $32924 \pm 14257$ & $14485 \pm 14718$ & $7828 \pm 394$ & $221473 \pm 105492$ & $13759 \pm 9673$ & $14656 \pm 312$ & $39645 \pm 24741$ & $53681 \pm 36092$ & $4035 \pm 163$ \\
                     & MAE  & $\mathbf{2244 \pm 275}$ & $4622 \pm 434$ & $23800 \pm 2048$ & $13720 \pm 55$ & $31243 \pm 1495$ & $32987 \pm 14202$ & $13609 \pm 13457$ & $7385 \pm 349$ & $203592 \pm 549573$ & $20729 \pm 8846$ & $14276 \pm 256$ & $39248 \pm 24898$ & $40897 \pm 27036$ & $4232 \pm 155$ \\ \midrule
\multirow{2}{*}{D14} & RMSE & $3929 \pm 407$ & $12393 \pm 124$ & $17807 \pm 407$ & $18383 \pm 75$ & $61838 \pm 3800$ & $12057 \pm 294$ & $\star \mathbf{2965 \pm 266}$ & $6293 \pm 214$ & $378793 \pm 13596$ & $96574 \pm 244$ & $10957 \pm 586$ & $63859 \pm 3939$ & $25569 \pm 1096$ & $13344 \pm 368$ \\
                     & MAE  & $\mathbf{4860 \pm 370}$ & $13205 \pm 2474$ & $15846 \pm 2996$ & $16410 \pm 67$ & $63807 \pm 3907$ & $12001 \pm 408$ & $6095 \pm 527$ & $5861 \pm 188$ & $378789 \pm 49343$ & $13112 \pm 1078$ & $10733 \pm 551$ & $60679 \pm 3933$ & $8135 \pm 753$ & $13346 \pm 373$ \\ \midrule
\multirow{2}{*}{D21} & RMSE & $\mathbf{94 \pm 8}$ & $17809 \pm 3462$ & $16956 \pm 665$ & $14895 \pm 58$ & $49345 \pm 2057$ & $41356 \pm 1350$ & $19040 \pm 7814$ & $11360 \pm 568$ & $91768 \pm 31601$ & $22011 \pm 1527$ & $22013 \pm 825$ & $44491 \pm 1680$ & $34724 \pm 2024$ & $251 \pm 7$ \\
                     & MAE  & $\mathbf{100 \pm 10}$ & $12152 \pm 2783$ & $16956 \pm 665$ & $13360 \pm 52$ & $67222 \pm 2786$ & $48950 \pm 1589$ & $22454 \pm 1089$ & $7760 \pm 399$ & $73038 \pm 13847$ & $27253 \pm 2499$ & $27135 \pm 485$ & $43176 \pm 1480$ & $33430 \pm 3433$ & $324 \pm 9$ \\ \midrule
\multirow{2}{*}{D22} & RMSE & $\mathbf{262 \pm 18}$ & $95945 \pm 3771$ & $18400 \pm 1940$ & $20686 \pm 81$ & $33049 \pm 663$ & $40445 \pm 4882$ & $11174 \pm 690$ & $10364 \pm 80$ & $118219 \pm 17533$ & $9329 \pm 712$ & $21742 \pm 250$ & $15115 \pm 1582$ & $53970 \pm 2637$ & $1985 \pm 68$ \\
                     & MAE  & $\mathbf{306 \pm 21}$ & $93949 \pm 4482$ & $18400 \pm 1940$ & $18560 \pm 73$ & $47993 \pm 657$ & $44622 \pm 5829$ & $11175 \pm 690$ & $9714 \pm 67$ & $111354 \pm 16089$ & $6831 \pm 598$ & $26178 \pm 171$ & $25413 \pm 1346$ & $50428 \pm 2477$ & $2040 \pm 88$ \\ \midrule
\multirow{2}{*}{D23} & RMSE & $\mathbf{2495 \pm 55}$ & $5413 \pm 236$ & $20593 \pm 5518$ & $60269 \pm 482$ & $28063 \pm 256$ & $24463 \pm 856$ & $18788 \pm 2000$ & $4357 \pm 236$ & $157454 \pm 21245$ & $16207 \pm 2146$ & $15685 \pm 1678$ & $52185 \pm 2253$ & $31013 \pm 1713$ & $4827 \pm 224$ \\
                     & MAE  & $\mathbf{2570 \pm 40}$ & $7424 \pm 277$ & $20593 \pm 5518$ & $53300 \pm 434$ & $36568 \pm 828$ & $24240 \pm 846$ & $10296 \pm 3381$ & $5849 \pm 314$ & $127321 \pm 25715$ & $12008 \pm 1418$ & $14106 \pm 1475$ & $52900 \pm 2558$ & $10762 \pm 1251$ & $7236 \pm 263$ \\ \midrule
\multirow{2}{*}{D24} & RMSE & $11343 \pm 431$ & $7076 \pm 278$ & $16219 \pm 1747$ & $67217 \pm 381$ & $25891 \pm 3905$ & $14973 \pm 3756$ & $18448 \pm 5630$ & $\star \mathbf{4537 \pm 564}$ & $750754 \pm 48253$ & $12154 \pm 1416$ & $8702 \pm 239$ & $15248 \pm 1280$ & $37149 \pm 3035$ & $28386 \pm 834$ \\
                     & MAE  & $11505 \pm 515$ & $9473 \pm 480$ & $16219 \pm 1748$ & $60495 \pm 343$ & $30163 \pm 4305$ & $14657 \pm 2531$ & $16421 \pm 5595$ & $\star \mathbf{4078 \pm 456}$ & $408696 \pm 14823$ & $13520 \pm 526$ & $16725 \pm 869$ & $17487 \pm 1056$ & $37487 \pm 3056$ & $29340 \pm 782$ \\ \midrule 
\multirow{2}{*}{D31} & RMSE & $\mathbf{43 \pm 15}$ & $111600 \pm 1453$ & $38899 \pm 529$ & $246 \pm 19$ & $595 \pm 8$ & $384 \pm 15$ & $10614 \pm 7085$ & $604 \pm 25$ & $31171 \pm 1912$ & $20527 \pm 4593$ & $122 \pm 27$ & $1624 \pm 631$ & $6254 \pm 2060$ & $908 \pm 47$ \\
                     & MAE  & $\mathbf{35 \pm 12}$ & $92628 \pm 1235$ & $32286 \pm 449$ & $201 \pm 15$ & $493 \pm 6$ & $318 \pm 12$ & $8809 \pm 5874$ & $501 \pm 21$ & $25871 \pm 1587$ & $17037 \pm 3812$ & $101 \pm 22$ & $1348 \pm 523$ & $5190 \pm 1710$ & $918 \pm 62$ \\ \midrule
\multirow{2}{*}{D32} & RMSE & $\mathbf{76 \pm 14}$ & $127962 \pm 2108$ & $39591 \pm 467$ & $268 \pm 18$ & $480 \pm 39$ & $397 \pm 25$ & $9859 \pm 216$ & $1093 \pm 76$ & $53452 \pm 2433$ & $24362 \pm 14002$ & $110 \pm 11$ & $2312 \pm 1154$ & $5660 \pm 1199$ & $957 \pm 64$ \\
                     & MAE  & $\mathbf{62 \pm 11}$ & $106208 \pm 1749$ & $32860 \pm 387$ & $222 \pm 15$ & $398 \pm 32$ & $329 \pm 20$ & $8183 \pm 179$ & $907 \pm 63$ & $44365 \pm 2019$ & $20221 \pm 11621$ & $91 \pm 9$ & $1919 \pm 958$ & $4697 \pm 995$ & $981 \pm 87$ \\ \midrule
\multirow{2}{*}{D33} & RMSE & $\mathbf{140 \pm 28}$ & $157826 \pm 3592$ & $38068 \pm 413$ & $335 \pm 78$ & $369 \pm 31$ & $433 \pm 34$ & $11490 \pm 913$ & $1242 \pm 28$ & $122171 \pm 18986$ & $20481 \pm 9290$ & $884 \pm 82$ & $13968 \pm 1242$ & $7843 \pm 724$ & $978 \pm 63$ \\
                     & MAE  & $\mathbf{115 \pm 23}$ & $131000 \pm 2981$ & $31596 \pm 342$ & $278 \pm 65$ & $306 \pm 26$ & $359 \pm 28$ & $9536 \pm 758$ & $1031 \pm 23$ & $101402 \pm 15758$ & $17001 \pm 7710$ & $733 \pm 68$ & $11593 \pm 1031$ & $6509 \pm 601$ & $978 \pm 53$ \\ \midrule
\multirow{2}{*}{D34} & RMSE & $\mathbf{243 \pm 33}$ & $138672 \pm 1988$ & $37938 \pm 386$ & $373 \pm 43$ & $334 \pm 139$ & $467 \pm 12$ & $7423 \pm 67$ & $834 \pm 14$ & $229863 \pm 67865$ & $56772 \pm 9475$ & $765 \pm 41$ & $22101 \pm 933$ & $4435 \pm 489$ & $1055 \pm 72$ \\
                     & MAE  & $\mathbf{199 \pm 27}$ & $115097 \pm 1650$ & $31488 \pm 320$ & $309 \pm 36$ & $277 \pm 115$ & $387 \pm 10$ & $6161 \pm 55$ & $692 \pm 12$ & $190786 \pm 56328$ & $47121 \pm 7864$ & $635 \pm 34$ & $18343 \pm 774$ & $3681 \pm 406$ & $1055 \pm 82$ \\ \midrule

\multicolumn{2}{l}{\textbf{Win/Loss}} & -- & 22/2 & 24/0 & 24/0 & 24/0 & 24/0 & 23/1 & 22/2 & 24/0 & 24/0 & 23/1 & 24/0 & 24/0 & 24/0 \\ \midrule
% 使用 \color{blue} 使整行数字变蓝
\multicolumn{2}{l}{\color{black}\textbf{Rank}} & \textbf{1.25} &8.25 & 9.08 & 6.12 & 9.25 & 7.75 & 6.96 & 4.54 & 13.21 & 7.71 & 6.25 & 9.67 & 10.12 & 4.83 \\ \midrule
\multicolumn{2}{l}{\textbf{\textit{p}-value}}  & -- & \textbf{1.67E-06} & \textbf{1.19E-07} & \textbf{1.19E-07} & \textbf{1.19E-07} & \textbf{1.19E-07} & \textbf{2.38E-07} & \textbf{2.78E-04} & \textbf{1.19E-07} & \textbf{1.19E-07} & \textbf{3.93E-06} & \textbf{1.19E-07} & \textbf{1.19E-07} & \textbf{1.19E-07} \\ \bottomrule
\end{tabular}
}
\end{table*}

\subsection{\textit{Speedup}}
\begin{figure*}[!htbp] 
    \centering
    % --- 子图 1 ---
    \begin{subfigure}[b]{0.32\linewidth} 
        \centering
        \includegraphics[width=\linewidth]{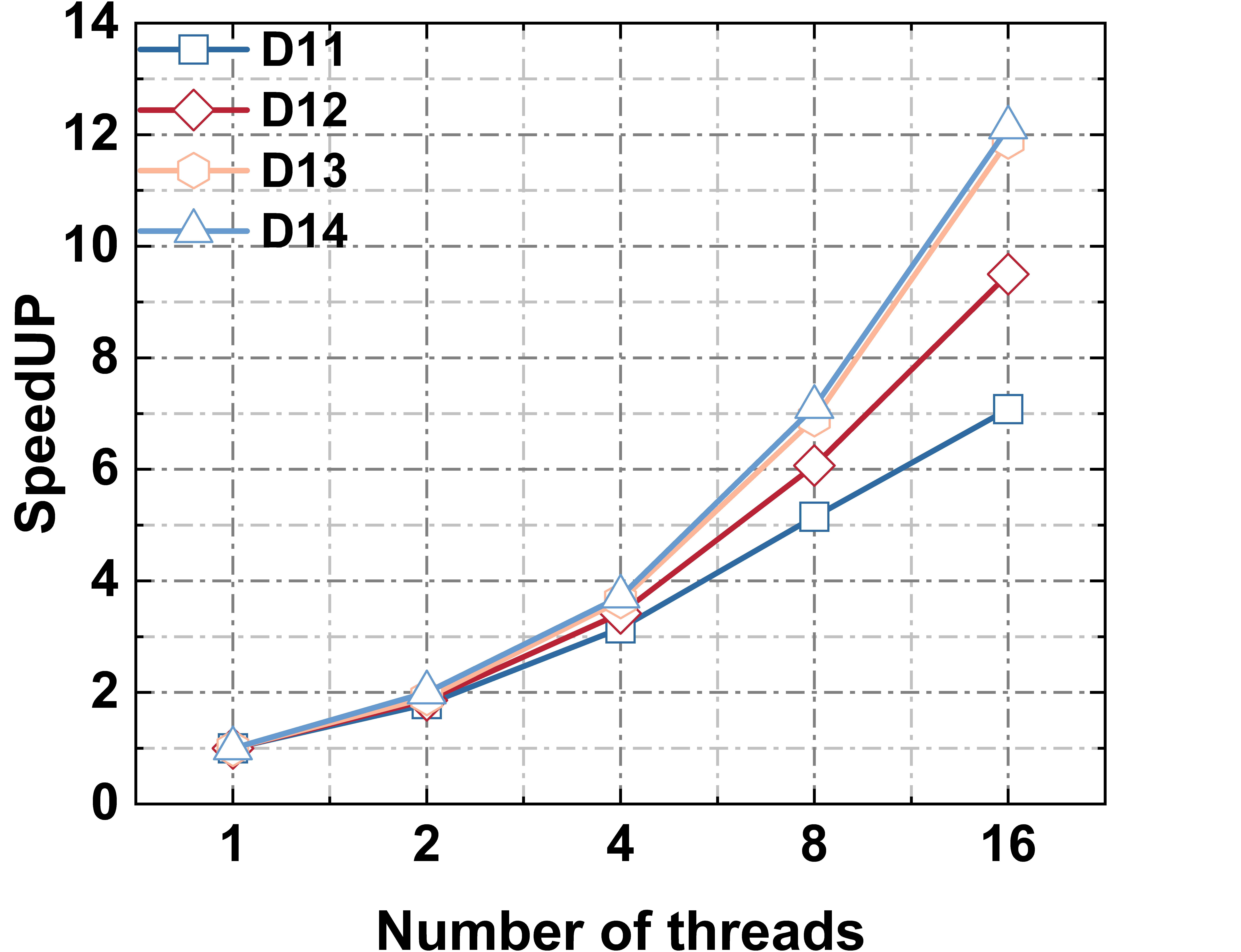} 
        \caption{}
        \label{fig:speedup_a}
    \end{subfigure}
    \hfill 
    % --- 子图 2 ---
    \begin{subfigure}[b]{0.32\linewidth}
        \centering
        \includegraphics[width=\linewidth]{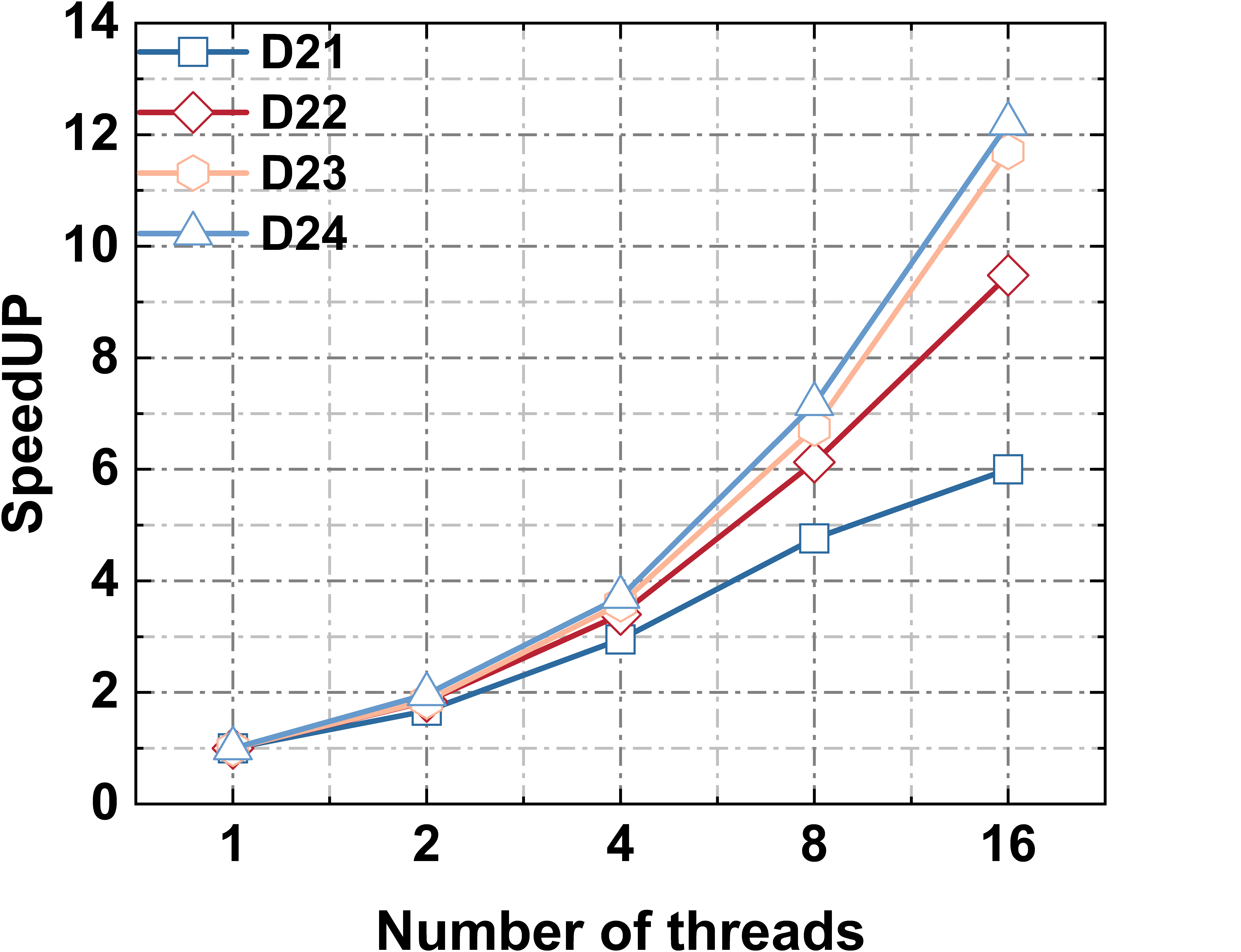} 
        \caption{}
        \label{fig:speedup_b}
    \end{subfigure}
    \hfill 
    % --- 子图 3 ---
    \begin{subfigure}[b]{0.32\linewidth}
        \centering
        \includegraphics[width=\linewidth]{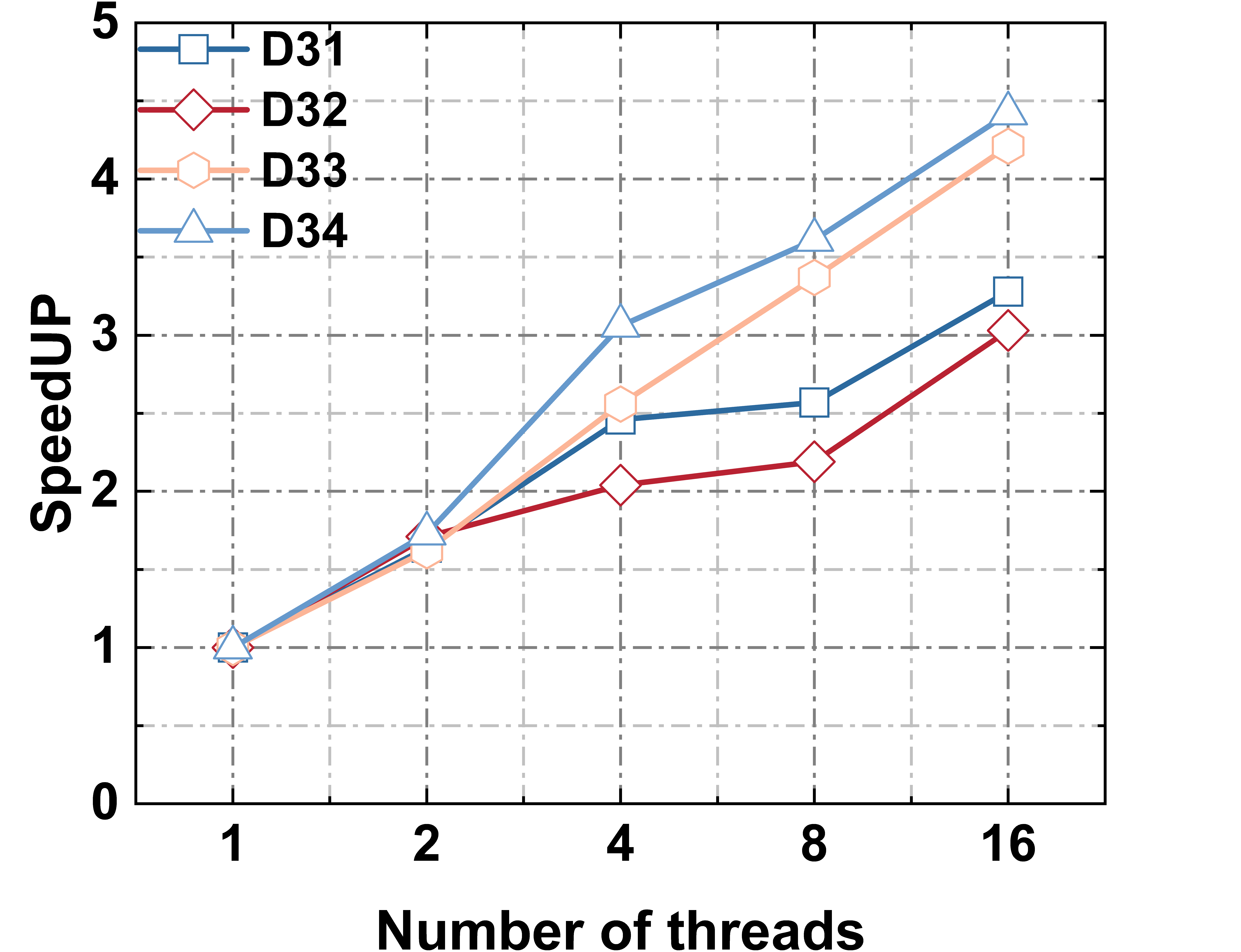} 
        \caption{}
        \label{fig:speedup_d3}
    \end{subfigure}
    
    \caption{Speedup of EKL as $M$ varies from 1 to 16 on D1-D3.}
    \label{fig:speedup_comparison}
\end{figure*}

We verify the effectiveness of parallel strategy in EKL. Firstly, we give the following formula to measure the speedup:
\[
\text{Speedup} = T_A / T_M,
\]
where $T_A$ and $T_M$ denote the total time costs of a serial version and its corresponding parallel implementation, respectively.

Note that the number of threads $M$ is $\{1, 2, 4, 8, 16\}$ in our context. The corresponding speedup is shown  in Fig. 6. From them, it is clear to see that EKL’s parallel efficiency is extremely remarkable. As shown in Fig. 6(b), on D23, the speedup reaches 11.70 with $M=16$, which indicates that the parallel implementation improves the computational efficiency by 11.7 times. We find that the speedup increases as the number of threads increases. For instance, on D23, the speedup is 1.85, 3.59, 6.74 and 11.70 when the value of $M$ is 2, 4, 8 and 16. This clearly demonstrates the positive impact of multi-thread parallel strategy on the efficiency. In addition, EKL shows the higher speedup on large-scale dataset. For instance, the speedup on D14 is 12.13. In contrast, the speedup on D11 is 7.08. The main reason is that the fixed parallel overheads, i.e., thread management and data splitting, take up a large share of total time. However, the substantial effective computation time on a large-scale dataset overshadows these overheads, which shows the advantages of parallel processing. However, the situation is different on D3, as depicted in Fig. 6(c). From it, we see that the speedup is significantly lower than the situations on D1 and D2. The reason lies in the extreme sparsity of D3, which leads to low proportion of effective computation. The additional overhead brought by thread management and data splitting cannot be covered by the computational efficiency gain. In summary, the proposed parallel strategy in this study can enhance EKL’s capability to process large-scale data in practical scenario applications.

\subsection{\textit{Convergence Study}}
To validate the convergence property of the proposed EKL, we conduct the corresponding convergence experiments on RMSE, as illustrated in Fig. 7 . From it, we find that EKL exhibits consistent and stable convergence behavior across all datasets. Specifically, RMSE declines sharply at the beginning stage of training and then tend to plateau, which indicates that the alternating optimization of temporal latent features via MFP and time-invariant latent features via DFP efficiently reduces the prediction loss. Notably, RMSE remains stable without divergence after convergence, which confirms that the alternating optimization strategy ensure the reliable convergence of EKL. It should be pointed out that EKL generally requires more training iterations on D3 compared with D1 and D2. It is probably because EKL extracts limited useful knowledge from the sparse dataset in each iterations. The convergence analysis experiment on MAE is presented in Fig. S1 of the Supplementary File.

\subsection{\textit{Hyperparameter Sensitivity}}
\begin{figure*}[!htbp]
    \centering
    % --- RMSE ---
    \begin{subfigure}[b]{0.32\linewidth}
        \centering
        \includegraphics[width=\linewidth]{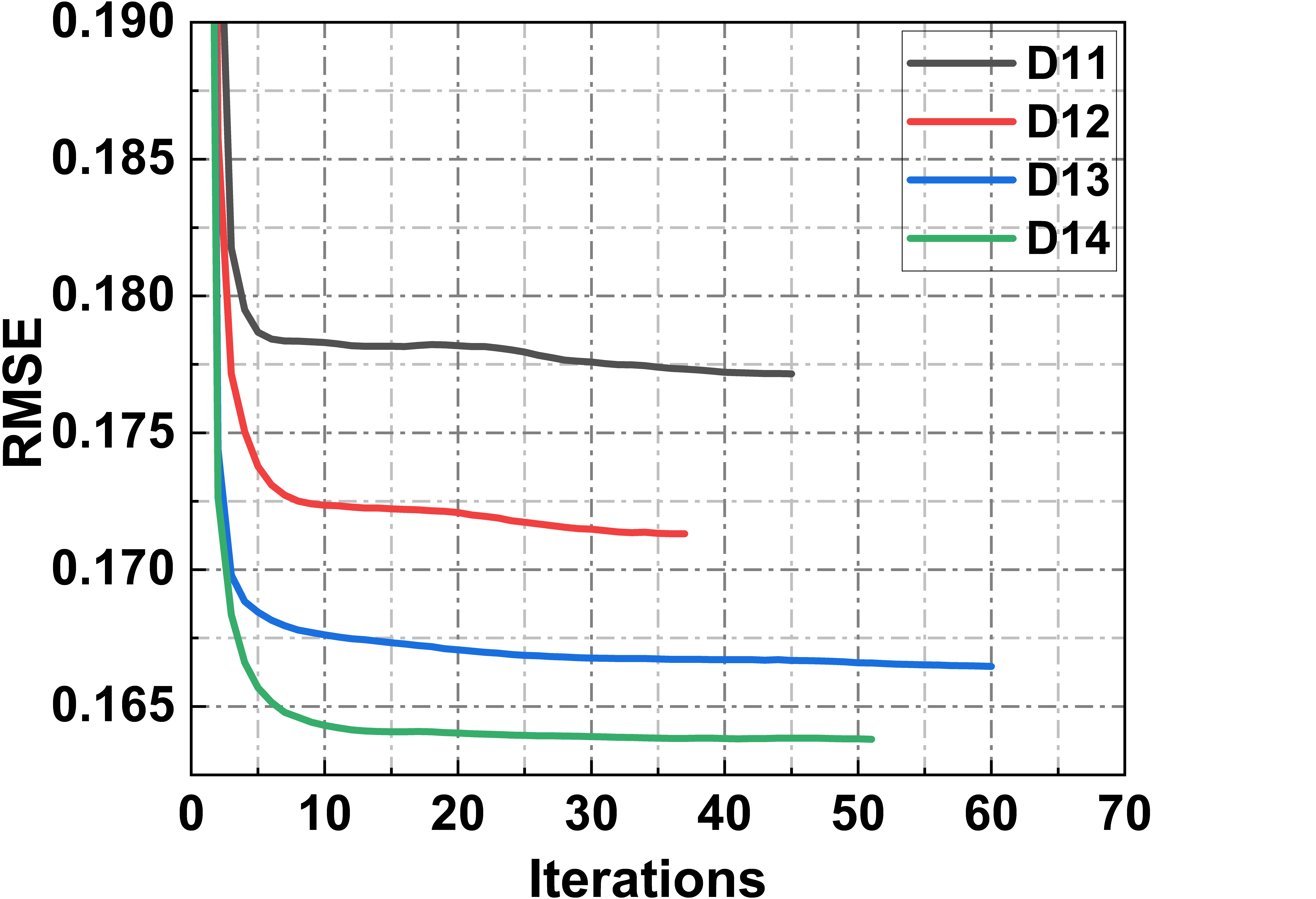}
        \caption{} 
        \label{fig:d1_rmse}
    \end{subfigure}
    \hfill 
    \begin{subfigure}[b]{0.32\linewidth}
        \centering
        \includegraphics[width=\linewidth]{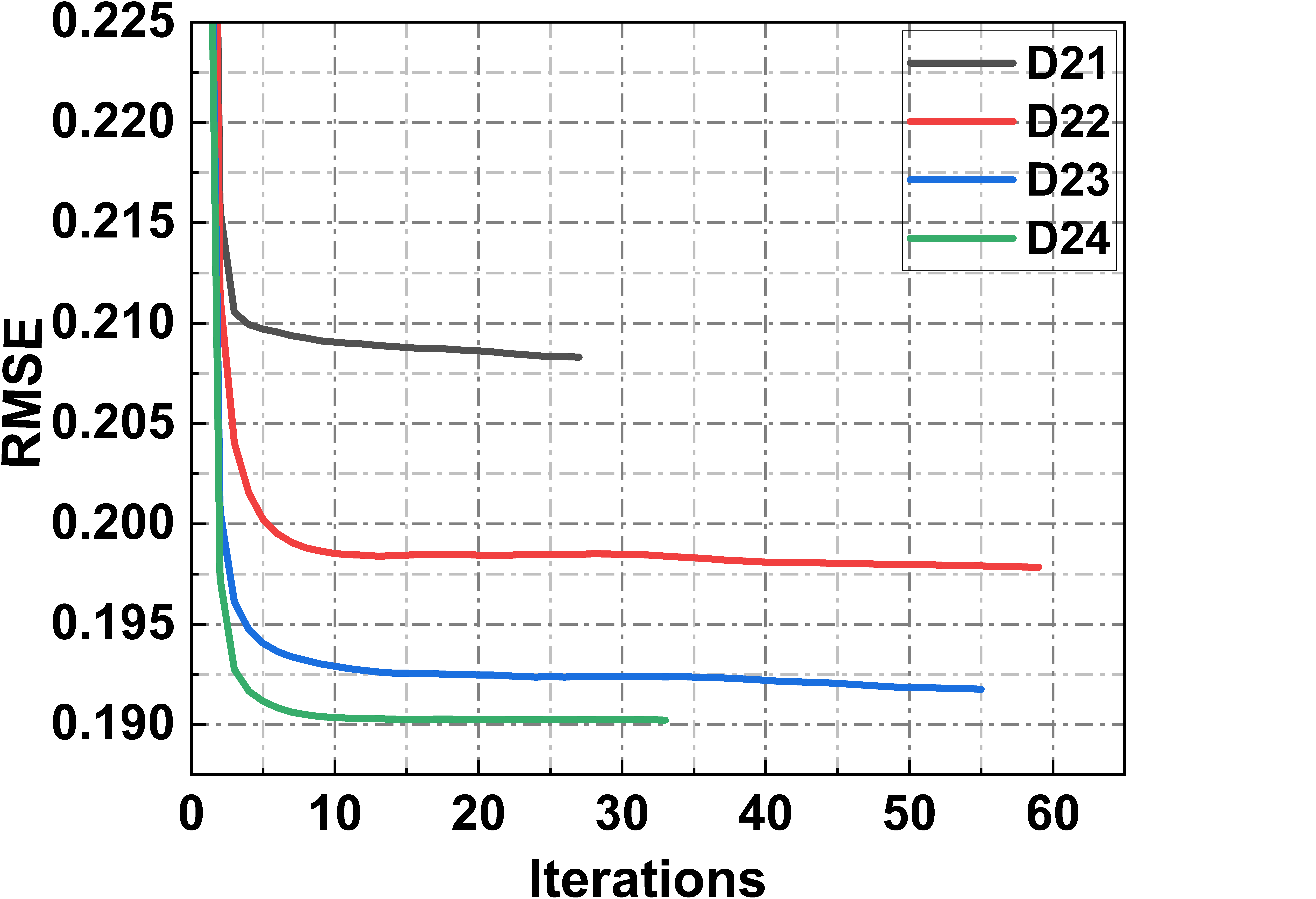}
        \caption{}
        \label{fig:d2_rmse}
    \end{subfigure}
    \hfill 
    \begin{subfigure}[b]{0.32\linewidth}
        \centering
        \includegraphics[width=\linewidth]{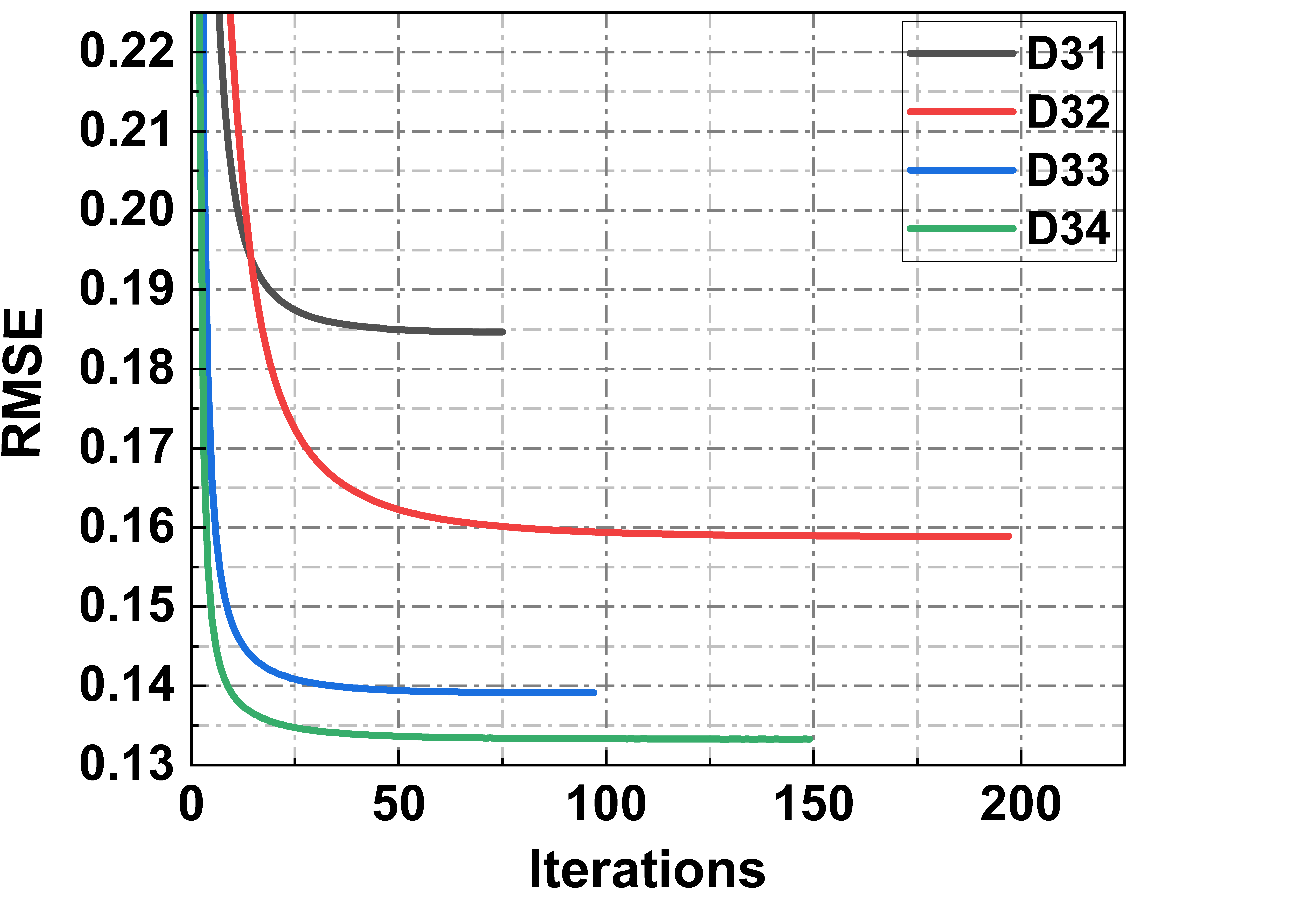}
        \caption{}
        \label{fig:d3_rmse}
    \end{subfigure}
    
    \caption{Convergence curves of RMSE on D1-D3 .}
    \label{fig:convergence_curves} 
\end{figure*}
It should be pointed out that EKL’s performance depends on some key hyperparameters like $\lambda$, $C[w_{(t)u}]$ and $C[r_{(t)u}]$. In addition, the covariance matrices $C[w_{(t)u}]$ and $C[r_{(t)u}]$ are simplified as $\sigma_w I$ and $\sigma_r I$ in our context. Hence, the corresponding tuning scales are $\lambda=\{ 0.001, \allowbreak 0.01, \allowbreak 0.1, \allowbreak 1, \allowbreak 10\}$, $\sigma_w=\{0.1, \allowbreak 1, \allowbreak 10, \allowbreak 100, \allowbreak 1000\}$, and $\sigma_r=\{\allowbreak 0.1, \allowbreak 1, \allowbreak 10, \allowbreak 100, \allowbreak 1000\}$. We utilize grid search to obtain the optimal hyperparameters of EKL based on the above scales. The optimal hyperparameters and total tuning time are summarized in Table~\ref{tab:optimal_hyperparams_detailed}. Fig. S2 of the Supplementary File depicts EKL’s performance as these hyperparameters vary on D11 and D21. Similar results are observed on other testing cases. From these results, it is clear that these hyperparameters need to be carefully tuned to ensure superior performance. For instance, as shown in Fig. S2(a) of the Supplementary File, the lowest RMSE value is 0.1773 when $\lambda=10^{-1}$, which is 9.26\% lower than the RMSE of 0.1954 when $\lambda=10^{-4}$. Considering the value of $\sigma_w$ and $\sigma_r$, they both significantly affect the estimation accuracy. For instance, the RMSE and MAE first decreases with the increase of $\sigma_w$ and $\sigma_r$, and then increases as the value of $\sigma_w$ and $\sigma_r$ exceeds a certain threshold. Moreover, a robust default setting and the corresponding estimation errors are provided to improve EKL' usability across datasets, as shown in Table~\ref{tab:optimal_hyperparams_detailed}. From it, we see that EKL still obtains competitive performance across datasets.

\begin{table}[!htbp]
    \centering
    \scriptsize % 缩小整个表格的字体
    \caption{The optimal hyperparameters and robust parameters on different cases.}
    \label{tab:optimal_hyperparams_detailed}
    \setlength{\tabcolsep}{3pt}
    \begin{tabular}{@{}llclcc@{}}
        \toprule
        \textbf{Cases} & \textbf{Optimal Setting} & \textbf{Time (hours)} & \textbf{Robust Setting} & \textbf{RMSE} & \textbf{MAE} \\ 
        \midrule
        D11 & $\lambda=0.1, \sigma_w=100, \sigma_r=1$ & 14.9 & \multirow{4}{*}{\begin{tabular}[c]{@{}l@{}}$\lambda=0.1$ \\ $\sigma_w=100$ \\ $\sigma_r=1$\end{tabular}} & 0.1786 & 0.1053 \\
        D12 & $\lambda=0.1, \sigma_w=100, \sigma_r=1$ & 26.3  & & 0.1717 & 0.1012 \\
        D13 & $\lambda=0.1, \sigma_w=100, \sigma_r=1$ & 103.3  & & 0.1667 & 0.0982 \\
        D14 & $\lambda=1, \sigma_w=100, \sigma_r=1$   & 213.4  & & 0.1652 & 0.0971 \\
        \midrule
        D21 & $\lambda=0.01, \sigma_w=100, \sigma_r=1$ & 9.9  & \multirow{4}{*}{\begin{tabular}[c]{@{}l@{}}$\lambda=0.1$ \\ $\sigma_w=100$ \\ $\sigma_r=1$\end{tabular}} & 0.2086 & 0.1175 \\
        D22 & $\lambda=0.1, \sigma_w=100, \sigma_r=1$  & 17.9  & & 0.1984 & 0.1111 \\
        D23 & $\lambda=1, \sigma_w=10, \sigma_r=1$     & 97.4  & & 0.1915 & 0.1072 \\
        D24 & $\lambda=1, \sigma_w=100, \sigma_r=1$    & 185.3  & & 0.1898 & 0.1068 \\
        \midrule
        D31 & $\lambda=0.1, \sigma_w=100, \sigma_r=1$    & 3.8  & \multirow{4}{*}{\begin{tabular}[c]{@{}l@{}}$\lambda=0.1$ \\ $\sigma_w=100$ \\ $\sigma_r=1$\end{tabular}} & 0.2047 & 0.0937 \\
        D32 & $\lambda=1, \sigma_w=1, \sigma_r=1000$   & 5.4  & & 0.1805 & 0.0828 \\
        D33 & $\lambda=1, \sigma_w=1, \sigma_r=100$    & 8.9  & & 0.1571 & 0.0719 \\
        D34 & $\lambda=0.1, \sigma_w=0.1, \sigma_r=1$  & 14.7  & & 0.1496 & 0.0697 \\ 
        \bottomrule
    \end{tabular}
\end{table}

\section{Conclusions}
This study propose a novel bidirectional model-data driven model named EKL for efficient and accurate QoS prediction, which addresses the limitations of existing purely data-driven QoS predictors. Specifically, EKL integrates a model-driven feature producer (MFP) based on the EKF to capture intricate temporal patterns of temporal QoS data. Further, a data-driven feature producer (DFP) utilizing ALS is designed to extract time-invariant latent features representing intrinsic user-service characteristics. Finally, a density-oriented parallel strategy (DPS) is adopted to significantly enhance computational efficiency. Furthermore, EKL can be conveniently integrated into an online autoscaler for predictive scaling to ensure SLO compliance. It delivers real-time, low-latency QoS predictions for unseen user–service invocations, which can be directly fed into the autoscaler to detect imminent SLO violations. Based on the results, cloud systems can proactively scale resources before performance degrades. Meanwhile, its DPS enables highly efficient inference, allowing EKL to meet the strict response-time requirements of online decision-making systems. In the future, we plan to do the following efforts:
\begin{enumerate}[label=\alph*)]
    \item EKL’s performance heavily depends on manual hyperparameter tuning, which is not only time-consuming but also lacks adaptive capability across different scenarios. To address this issue, we intend to incorporate Bayesian optimization \cite{kent2024bayesian, lin2026ncsac, yang2025link} or reinforcement learning \cite{zhang2025multi} to develop an adaptive hyperparameter tuning mechanism;
    \item EKL simplifies the state-transition and observation noises as Gaussian distributions. In real scenarios, noises within QoS data often exhibits spatial-temporal correlation. Hence, we aim to design a spatial-temporal correlation module like graph neural network layer \cite{liu2023qosgnn, gou2026multiscale, he2025structure, wu2025graphmirna} to capture noise correlations across users/services or time slots.
    \item We intend to investigate the performance of various nonlinear activation functions, e.g., GELU, Swish, Tanh, within the state-transition and observation functions of EKF, and further analyze the matching mechanism between these activation functions and the fluctuation characteristics of temporal QoS data \cite{yang2025fmvpci, deng2026fuzzy, liu2025sentiment}.
\end{enumerate}

\end{document}